%% file: main.tex
\documentclass[10pt,twocolumn,letterpaper]{article}

\usepackage[pagenumbers]{cvpr} 

\input{preamble}
\definecolor{cvprblue}{rgb}{0.21,0.49,0.74}
\usepackage[pagebackref,breaklinks,colorlinks,allcolors=cvprblue]{hyperref}
\usepackage{xspace}
\usepackage{balance}

\newcommand{\methodname}{V-Warper\xspace}
\newcommand{\ours}{\methodname}

\def\paperID{7162} 
\def\confName{CVPR}
\def\confYear{2026}

\title{\ours: Appearance-Consistent Video Diffusion Personalization \\ via Value Warping}

\author{
Hyunkoo Lee\textsuperscript{1} \qquad Wooseok Jang\textsuperscript{1} \qquad Jini Yang\textsuperscript{1} \qquad Taehwan Kim\textsuperscript{1} \\ \qquad Sangoh Kim\textsuperscript{1} \qquad Sangwon Jung\textsuperscript{1,2} \qquad Seungryong Kim\textsuperscript{1}$^{\dagger}$ \\[5pt]
 \textsuperscript{1}KAIST AI \qquad  \textsuperscript{2}Korea University \\[5pt]
{\tt \href{https://cvlab-kaist.github.io/V-Warper}{https://cvlab-kaist.github.io/V-Warper}}
}

\begin{document}


\twocolumn[{
    \maketitle
    \begin{center}
    \includegraphics[width=\textwidth]{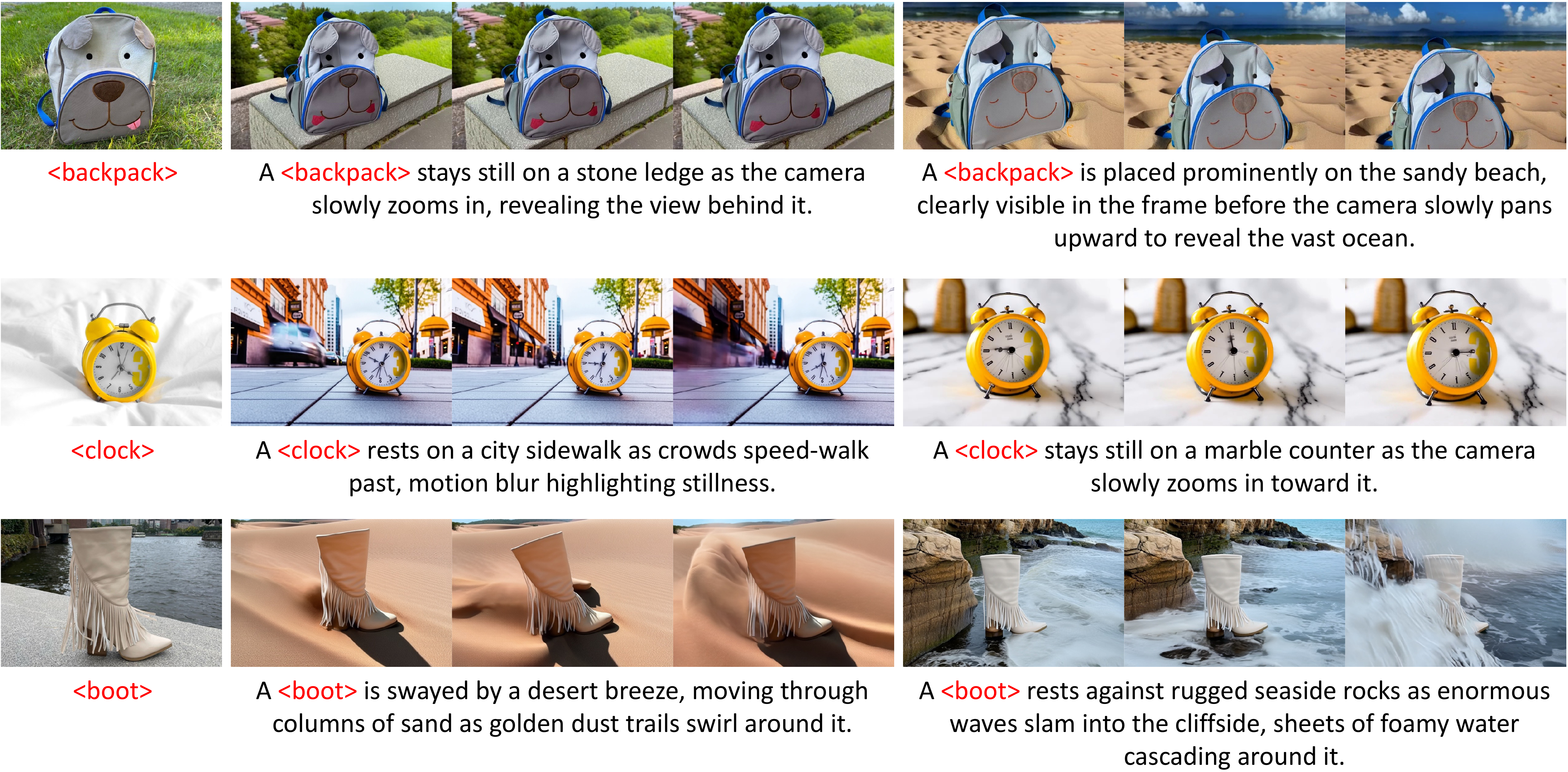}
    \begingroup
        \captionsetup{hypcap=false}
         \captionof{figure}{
             \textbf{Video diffusion personalization examples.}
             Given a few reference images, our training-free framework, called \textbf{\ours}, generates high-fidelity videos that accurately preserve the appearance of the reference subjects while following the provided text prompts.
         }
         \label{fig:teaser}
    \endgroup
    \end{center}
}]

\input{sec/0_abstract}    
\input{sec/1_intro}
\input{sec/2_related_works}
\input{sec/3_method}

\input{sec/4_experiment}
\input{sec/5_conclusion}

\input{sec/X_suppl}

\clearpage
\balance
{
    \small
    \bibliographystyle{ieeenat_fullname}
    \bibliography{main}
}


\end{document}

%% file: sec/0_abstract.tex
\begin{abstract}
Video personalization aims to generate videos that faithfully reflect a user-provided subject while following a text prompt.
However, existing approaches often rely on heavy video-based finetuning or large-scale video datasets, which impose substantial computational cost and are difficult to scale. Furthermore, they still struggle to maintain fine-grained appearance consistency across frames.
To address these limitations, we introduce \textbf{\ours}, a training-free coarse-to-fine personalization framework for transformer-based video diffusion models. The framework enhances fine-grained identity fidelity without requiring any additional video training.
(1) A lightweight \textit{coarse appearance adaptation} stage leverages only a small set of reference images, which are already required for the task. This step encodes global subject identity through image-only LoRA and subject-embedding adaptation.
(2) A inference-time \textit{fine appearance injection} stage refines visual fidelity by computing semantic correspondences from RoPE-free mid-layer query--key features. These correspondences guide the warping of appearance-rich value representations into semantically aligned regions of the generation process, with masking ensuring spatial reliability.
\ours significantly improves appearance fidelity while preserving prompt alignment and motion dynamics, and it achieves these gains efficiently without large-scale video finetuning.
\end{abstract}

%% file: sec/1_intro.tex
\begin{figure*}[t]
    \centering
    \includegraphics[width=\textwidth]{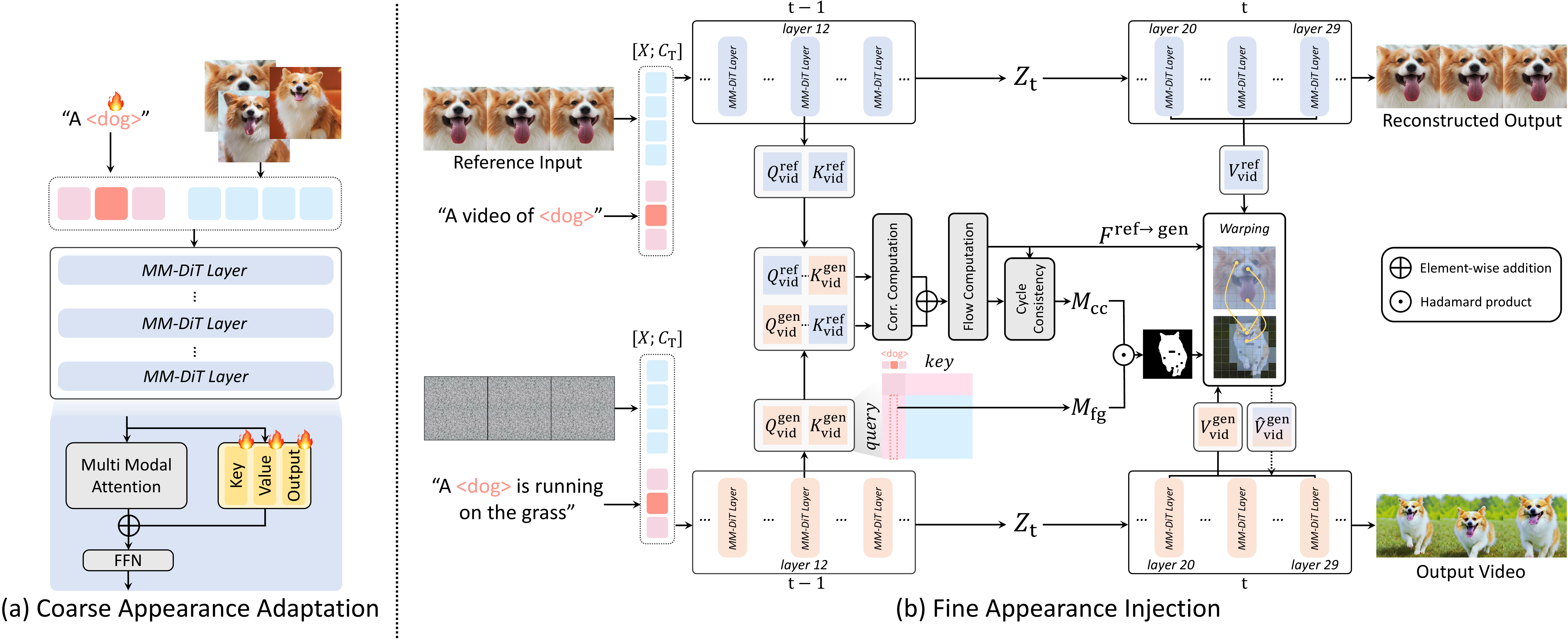}
    \caption{\textbf{Overall architecture of \ours.}
    (a) A lightweight coarse appearance adaptation stage trains LoRA modules and a learnable subject token using only a small set of reference images, enabling the model to encode global subject identity.
    (b) A inference-time fine appearance injection stage computes semantic correspondences from RoPE-free mid-level query--key features and warps appearance-rich value features into semantically matched regions of the generation process, reinforced by masking for spatial reliability.
    }
    \label{fig:figure2_architecture}
\end{figure*}

\vspace{-10pt}
\section{Introduction}
\label{sec:intro}

Text-to-video (T2V) diffusion models~\cite{yang2024cogvideox, kong2024hunyuanvideo, blattmann2023stablevd, wan2025wan, chen2025goku, hacohen2024ltx, ho2022imagenvideo, wang2023modelscopetexttovideotechnicalreport}
have made remarkable progress in generating high-quality and temporally coherent videos from textual prompts.
With these advances, \textit{video personalization} has emerged as an important direction for controllable video generation. It aims to create videos that faithfully depict a user-provided subject while following text instructions.
This capability enables diverse applications such as personalized content creation~\cite{wei2023dreamvideo, jiang2024videobooth, huang2025videomage, kim2025subject}, virtual avatar generation~\cite{yang2024amg, yu2025llia}, and cinematic production~\cite{xing2025motioncanvas, lei2025animateanything}.

Personalization has been studied extensively in the image domain through 
optimization-based fine-tuning, task-specific supervised training on additional datasets, and 
training-free attention manipulation~\cite{ruiz2023dreambooth, gal2022textual, shi2023masactrl, ye2023ip, nam2024dreammatcher, tan2025ominicontrol}.
However, extending image techniques to video remains challenging.
Video diffusion models must preserve temporal coherence and maintain motion priors learned from large-scale video training.
Directly applying image-based personalization often disrupts these priors, leading to identity drift or motion instability.

Recent video personalization approaches attempt to overcome this limitation from two directions.
Some approaches modify only spatial components within architectures that decouple spatial and temporal pathways~\cite{wei2023dreamvideo}. However, this strategy does not naturally extend to recent 3D full-attention T2V models, where spatial and temporal information are tightly coupled.
Other methods incorporate auxiliary video supervision or temporal regularization~\cite{huang2025videomage, kim2025subject}. Nevertheless, they rely on heavy fine-tuning and large task-specific video datasets, which are computationally expensive and difficult to scale.
Despite these efforts, both categories still struggle to preserve fine-grained appearance across frames and often exhibit identity drift or temporal inconsistencies.

To overcome these limitations, we explore an alternative direction that strengthens fine-grained subject appearance without relying on additional video data or large-scale training.
We focus on injecting only appearance-related information so that the model can enrich subject details without disturbing its learned temporal prior.
To make this enhancement accurate, the injected information must be placed only at spatially appropriate and semantically corresponding regions of the video.
To support such precise control, we analyze the internal representations of video diffusion transformers and observe meaningful semantic alignment, consistent with patterns reported in image diffusion models.
This finding motivates transferring appearance-rich features to semantically matched regions, as in DreamMatcher~\cite{nam2024dreammatcher}, enabling precise and training-free appearance enhancement.

Building on these observations, \textbf{\ours} adopts a coarse-to-fine personalization strategy, as illustrated in Fig.~\ref{fig:figure2_architecture}.
First, a lightweight \textbf{coarse appearance adaptation} stage uses only a few reference images, which are already required for the task, to fit LoRA modules and a learnable subject token. This step captures global subject identity with minimal overhead while remaining independent of any video data.
Next, a inference-time \textbf{fine appearance injection} stage computes semantic correspondences using RoPE-free mid-layer query--key features and uses them to warp appearance-rich value features into the generation process. Masking restricts this injection to reliable, subject-relevant regions, ensuring that fine-grained details are enhanced only where semantically appropriate.


Our contributions are summarized as follows:
\begin{itemize}
    \item We introduce \textbf{\ours}, a training-free coarse-to-fine video personalization framework that enhances fine-grained appearance fidelity without relying on large-scale video data.
    \item We provide a \textbf{systematic analysis of semantic correspondences} in video diffusion transformers and show that RoPE-free mid-layer query--key features yield the most reliable alignment for appearance transfer.
    \item We demonstrate that \ours achieves \textbf{state-of-the-art appearance fidelity} while preserving text alignment and motion dynamics, validated through quantitative metrics and human evaluations.
\end{itemize}

%% file: sec/2_related_works.tex
\section{Related Work}
\label{sec:related}

\paragraph{Image Personalization.}
Personalization methods for text-to-image diffusion models~\cite{NEURIPS2020_ddpm,podell2023sdxlimprovinglatentdiffusion_sdxl,Rombach_2022_CVPR_ldm} can be grouped into three categories depending on whether they require subject-specific optimization, method-level training, or no additional training. Optimization-based methods~\cite{gal2022textual,ruiz2023dreambooth,Kumari_2023_CVPR_customdiffusion} personalize a model through instance-specific optimization, typically by fine-tuning model parameters or learning new subject embeddings. Training-based methods~\cite{ye2023ip,tan2025ominicontrol} train auxiliary modules that enable feed-forward personalization without user-specific tuning, such as adapter-based image conditioning or lightweight control pathways for DiT models. Training-free methods~\cite{nam2024dreammatcher, shin2025large, feng2025personalize} perform personalization purely at inference time by manipulating attention or feature pathways, leveraging semantic correspondence, structured prompting, or token-level feature injection.


\paragraph{Video Personalization.}
Personalizing video generation requires preserving both temporal coherence and subject identity. 
Early approaches address these challenges through model adaptation.
VideoBooth~\cite{jiang2024videobooth} learns a coarse subject embedding with a frozen CLIP encoder and then refines appearance by fine-tuning the diffusion model’s K/V projections. 
DreamVideo~\cite{wei2023dreamvideo} instead adopts an optimization-driven design, using its separated spatial and temporal transformers to train independent ID and motion adapters.
More recent systems broaden controllability and subject configurations. 
VACE~\cite{jiang2025vace} unifies reference-to-video generation and multiple editing tasks within one framework. 
VideoMage~\cite{huang2025videomage} handles multi-subject personalization and their interactions. 
HunyuanCustom~\cite{hu2025hunyuancustom} extends customization to multi-modal inputs, including images, audio, video, and text. 
A parallel line of work~\cite{kim2025subject} disentangles identity from motion by injecting subject information while preserving temporal dynamics through stochastically switching over video data.
Unlike these approaches, which rely on specialized adapters or fine-tuning, our method applies training-free feature warping to a DiT-based video diffusion model, enabling effective and flexible appearance transfer.

%% file: sec/3_method.tex
\begin{figure}[t]
    \centering
    \includegraphics[width=1.00\linewidth]{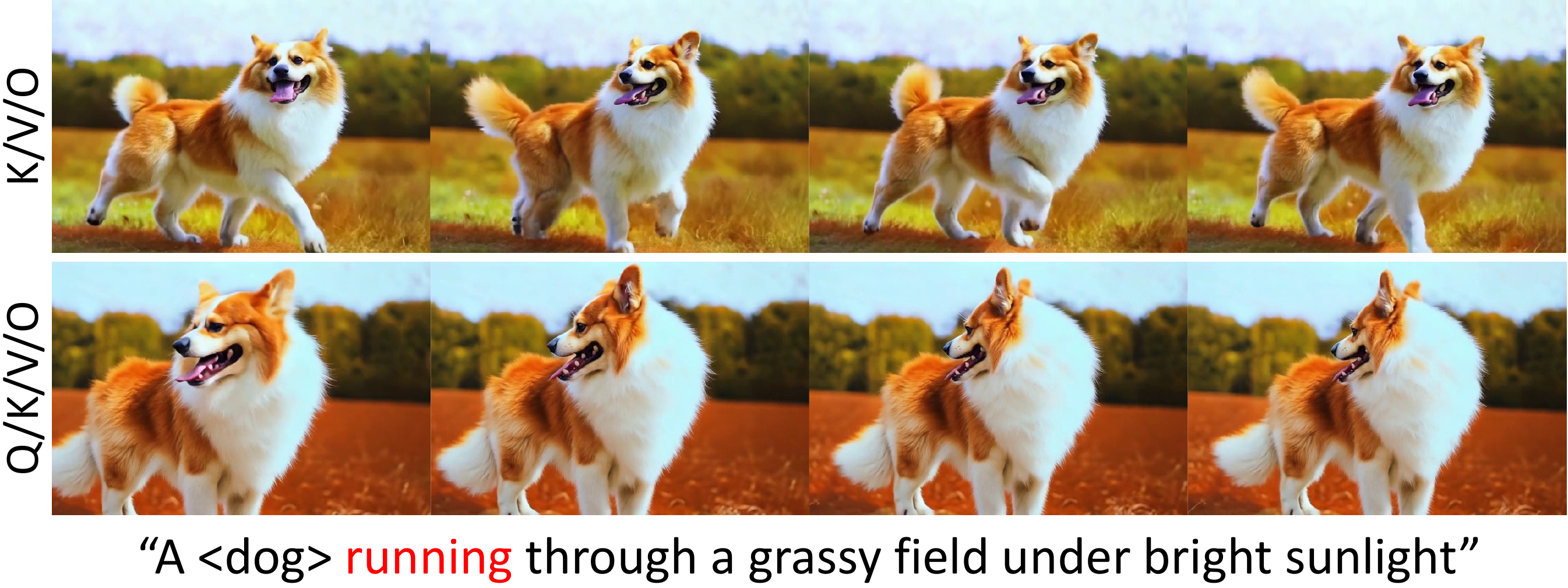}
    \caption{
\textbf{Comparison of motion alignment across different updated attention components.}
Freezing the query (Q) and updating only the key (K), value (V), and output projection (O) yields motion that better follows the text prompt.}
    \label{fig:lora_tgt_module}
\end{figure}

\section{Preliminaries}
\label{sec:preliminaries}

\paragraph{Video diffusion models.}
VDMs generate videos by progressively denoising a sequence of noisy samples drawn from a Gaussian distribution~\cite{NEURIPS2020_ddpm}.
In modern text-to-video systems, the diffusion process operates in a VAE-compressed latent space~\cite{kingma2022autoencodingvariationalbayes_vae}.
At each timestep $t$, the model receives a noisy video latent $x_t$ and predicts the noise component $\epsilon$ that was added during the forward process.
A neural network $\epsilon_\theta(x_t, t, c)$, conditioned on a text prompt $c$, is trained to approximate this noise through a simple reconstruction objective:
\begin{equation}
\mathcal{L}
= 
\mathbb{E}_{x, \epsilon, t}
\!\left[
\|
\epsilon_\theta(x_t, c, t) - \epsilon
\|_2^2
\right].
\end{equation}
Starting from Gaussian noise $x_T \!\sim\! \mathcal{N}(\mathbf{0},\mathbf{I})$,  
the trained model iteratively removes noise across timesteps to generate a coherent video sequence $x_0$ consistent with the conditioning signal $c$.

\paragraph{Multi-Modal Diffusion Transformers.}
Multi-Modal Diffusion Transformers (MM-DiT)~\cite{tan2025ominicontrol, kim2025subject,yang2024cogvideox,kong2024hunyuanvideo,blackforestlabs2024flux} extend the DiT~\cite{peebles2023scalable} architecture to process text and visual modalities within a unified Transformer framework.  
Positional information is encoded explicitly via Rotary Position Embeddings (RoPE)~\cite{su2024roformer} and applied to visual tokens.

The noisy video latent $x_t$ is patchified into visual tokens $X \in \mathbb{R}^{N \times d}$, and the text prompt $c$ is encoded into text tokens $C_T \in \mathbb{R}^{M \times d}$.  
These tokens are concatenated into a single sequence $[X; C_T]$, which is jointly processed by the Multi-Modal Attention (MMA) mechanism:
\begin{equation}
\mathrm{MMA}([X;C_T]) = \mathrm{softmax}\!\left(\frac{QK^{\top}}{\sqrt{d}}\right)V.
\label{eq:mma}
\end{equation}
Through this unified attention, text and visual tokens exchange information bidirectionally, enabling coherent multi-modal conditioning during denoising.

\section{Method}

Existing video personalization approaches typically rely on video-based finetuning or auxiliary temporal supervision to preserve motion priors~\cite{huang2025videomage, kim2025subject}.  
However, these strategies still lack fine control over where appearance information is injected, often causing identity drift or structural distortions.

To overcome this limitation, we introduce \textbf{\ours}, a training-free coarse-to-fine framework for subject-driven video personalization. 
The method consists of two stages: 
(1) \textbf{Coarse Appearance Adaptation}, which performs a minimal image-only adjustment using LoRA and a learnable subject embedding to encode global subject identity;
and (2) \textbf{Fine Appearance Injection}, which transfers fine-grained appearance by computing token-level correspondences between a reference and generation branch and warping appearance-rich value features with masking.

This two-stage design enables high-fidelity identity preservation without large-scale video training and applies broadly to MM-DiT based video diffusion models.

\subsection{Coarse Appearance Adaptation}
\label{subsec:coarse}

The first stage of \ours adapts the video diffusion model to the subject’s coarse appearance using a few reference images. 
As illustrated in Fig.~\ref{fig:figure2_architecture} (left), this stage introduces a learnable subject token with a trainable text embedding that encodes coarse subject characteristics in the text embedding space.  
This embedding is optimized jointly with LoRA modules attached to the attention layers of the MM-DiT backbone, enabling efficient adaptation without full fine-tuning.

However, training solely on static images introduces a modality gap that can distort the video model’s pretrained temporal behavior.
To mitigate this issue, we avoid updating the query projection, as it encodes much of the model’s temporal structure \cite{atzmon2024motion}.  
Instead, we apply LoRA only to the key, value, and output projections.
As shown in Fig.~\ref{fig:lora_tgt_module}, updating all query (Q), key (K), value (V), and output (O) projections degrades temporal dynamics and can suppress natural motion, turning a running subject into a nearly static one. 
In contrast, restricting LoRA to only K/V/O preserves expected motion while still incorporating appearance-related features.
Motivated by this observation, we adopt K/V/O-only LoRA as a lightweight and stable adaptation strategy.

While this stage successfully establishes coarse subject identity, it still fails to capture fine-grained details, such as subtle texture and color nuances that are crucial for visual realism.
To address this limitation, we introduce a subsequent refinement stage that injects high-frequency appearance information directly from the reference image. 

\begin{figure}[t]
    \centering
    \includegraphics[width=\columnwidth]{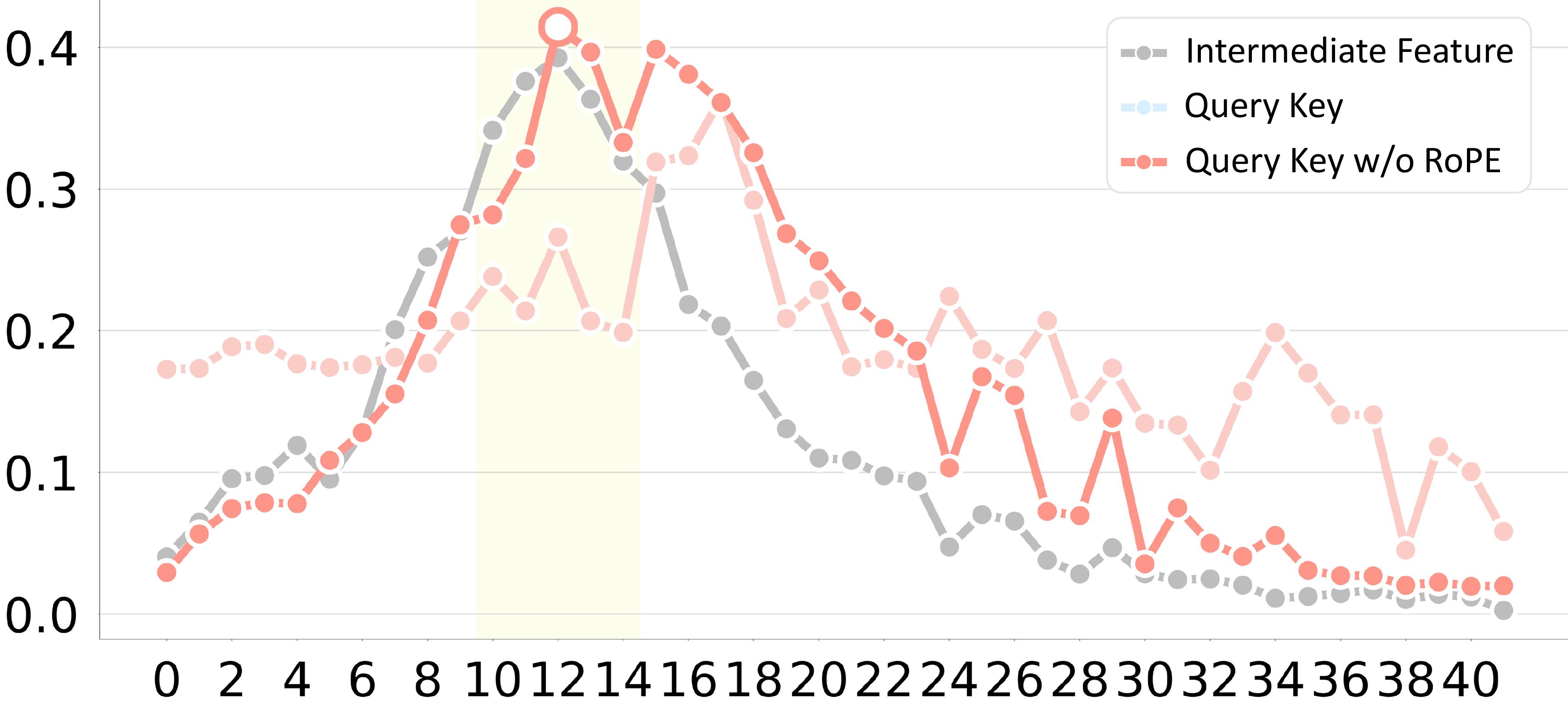}
    \caption{\textbf{Quantitative semantic matching performance.}
We compute PCK accuracy across all 42 layers of CogVideoX-5B for three 
candidate descriptor types.  
RoPE-free query--key features at mid-level layers show the strongest semantic 
alignment, with Layer~12 achieving the highest accuracy.  
}
    \label{fig:pck_layer}
\end{figure}

\subsection{Emergent Semantic Correspondence Analysis}
\label{subsec:semantic_analysis}

Our fine appearance injection module requires accurate token-level semantic 
correspondences between the reference and generation branches.  
While prior work such as DiffTrack~\cite{nam2025emergent} has shown that
diffusion transformers implicitly learn temporal correspondences across video frames, it remains unclear whether they also encode cross-branch semantic alignment. This capability is essential for transferring appearance information in video personalization.

To this end, we evaluate several internal feature types from the 
video DiT backbone as potential semantic matching descriptors.  
First, \textit{intermediate features} are included as a baseline candidate, 
as they contain mixed spatial and textual information and have been shown to 
support dense geometric or semantic matching~\cite{NEURIPS2023_dift,NEURIPS2023_sd_dino,nam2024diffusionmodeldensematching,li2024sd4matchlearningpromptstable}.
Second, \textit{query--key attention features} are considered because 
attention in diffusion models has been empirically shown to encode object-level 
semantics and structural cues~\cite{liu2024understandingcrossselfattentionstable,nam2025emergent}. 
Finally, we also examine \textit{query--key features without RoPE}.  
Rotary position embeddings are known to introduce strong location-dependent bias, which disrupts accurate semantic matching by encouraging same-location interactions~\cite{feng2025personalize,tan2025ominicontrol}.
Removing RoPE therefore allows us to evaluate whether the model encodes 
position-invariant semantic similarity that is suitable for appearance transfer.

To identify which feature type and layer provide the most reliable semantic alignment, we quantitatively assess correspondence accuracy across all layers of CogVideoX-5B using PCK~\cite{yang2012articulated} measured over subject foreground regions.
Dense correspondences computed by DINOv3~\cite{simeoni2025dinov3} on generated frames serve as pseudo ground truth.
The results, summarized in Fig.~\ref{fig:pck_layer}, reveal two consistent 
patterns.  
First, \textbf{RoPE-free query--key features outperform both intermediate 
features and raw query--key features} across most of the layers, demonstrating that removing RoPE significantly reduces positional bias and enhances semantic alignment.  
Second, \textbf{mid-level layers (10--20)} provide the strongest correspondence performance overall, with \textbf{Layer~12} achieving the highest PCK score.  
These findings establish Layer~12 RoPE-free query--key features as the most 
reliable semantic descriptor for our matching module.
Further details of the evaluation protocol are provided in Appendix~\ref{sec:matching}.

\begin{figure}[t]
    \centering
    \includegraphics[width=1.0\linewidth]{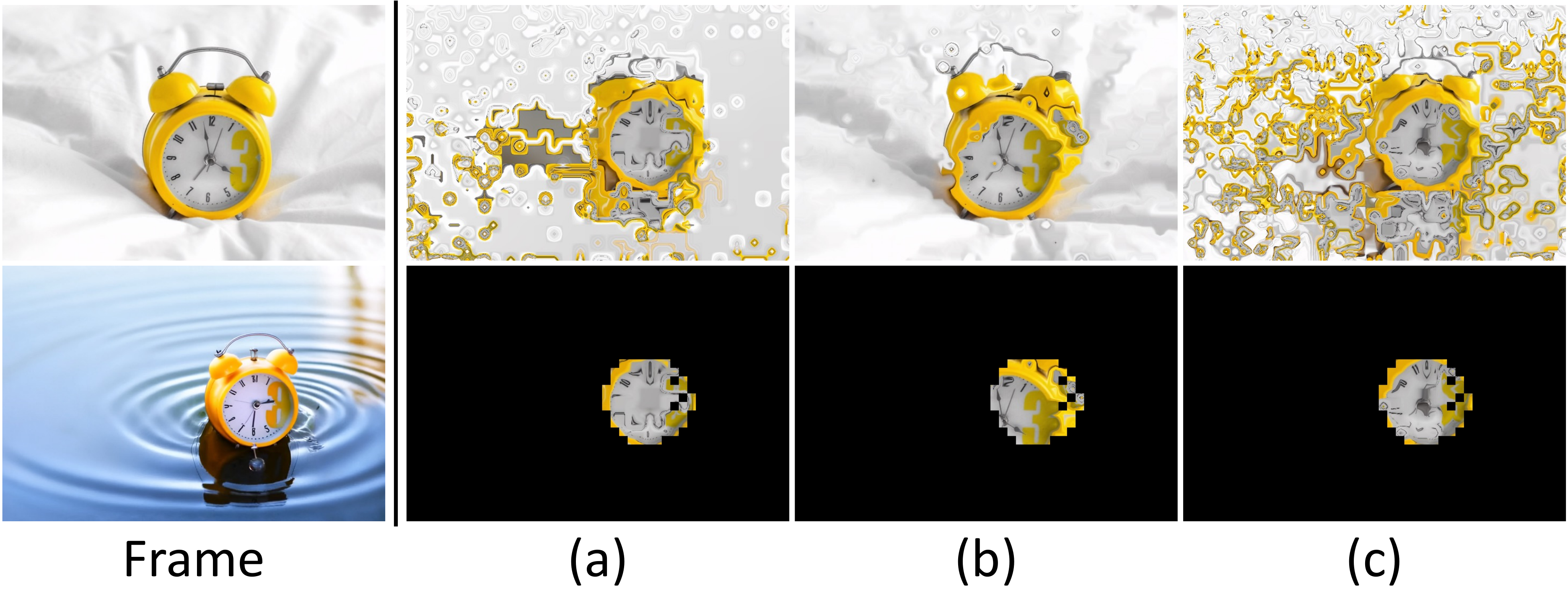}
    \caption{\textbf{Qualitative semantic matching performance.}
Left: reference (top) and generation (bottom) frames.  
Right: reference-frame pixels warped into the target frame using correspondences from each feature type  
(top: unmasked results, bottom: masked results).  
(a) Intermediate features yield weak alignment.  
(b) Raw query--key features exhibit strong RoPE-induced positional bias.  
(c) RoPE-free query--key features provide the most accurate semantic matches.}
    \label{fig:correspondence_vis}
\end{figure}

To further validate the quantitative trends, we visualize cross-branch
correspondences for different feature types.
This visualization is obtained by warping pixels from the reference frame
into the generation frame using the estimated correspondence map.
As shown in Fig.~\ref{fig:correspondence_vis}, \textit{intermediate features} 
produce weak and spatially inconsistent swaps, failing to transfer fine details 
such as the internal digits of the clock or the thin connector between the bells.  
\textit{Raw query–key features} also perform poorly due to strong
RoPE-induced positional bias. This bias pulls warped pixels toward their original spatial coordinates, causing the subject to reappear near the reference location instead of its true position in the target frame.
In contrast, \textit{RoPE-free query--key features} yield clean, structurally 
faithful correspondences that correctly map semantic parts to the appropriate 
locations in the generation frame.  This qualitative behavior aligns precisely 
with the quantitative PCK results.

Together, these analyses demonstrate that video DiT models not only learn 
temporal coherence but also encode cross-branch semantic alignment suitable for 
appearance transfer.  
Based on the combined quantitative and qualitative evidence, \ours adopts 
\textbf{RoPE-free Layer~12 query--key features} as the matching representation 
for all subsequent appearance injection operations.

\subsection{Fine Appearance Injection}
\label{subsec:fine}

Having identified a reliable semantic matching representation in
Sec.~\ref{subsec:semantic_analysis}, we now describe how it is used to
inject fine-grained appearance into the video generation process.
Whereas the coarse adaptation stage provides a subject-aware prior,
the fine stage enriches the generation with high-frequency appearance details drawn directly from the reference image.

As illustrated in Fig.~\ref{fig:figure2_architecture} (right),
we form two synchronized branches during inference. 
One branch is a reference branch obtained by inverting the reference image,
and the other is a generation branch driven by the text prompt.
Cross-branch semantic correspondences computed during denoising guide a
feature-warping module that transfers appearance-rich value features from
the reference branch into the generation branch, improving subject fidelity
without modifying the diffusion model.

\paragraph{Correlation Computation.}
At each denoising step $t$, token-level correspondences are computed using 
the Layer~12 query--key representations from the previous step after removing RoPE.  
These representations serve as the matching descriptor identified in 
Sec.~\ref{subsec:semantic_analysis}.  
Given the resulting query and key features 
$\bar{Q}^{(\cdot)}_{t-1}$ and $\bar{K}^{(\cdot)}_{t-1}$ from 
the reference and generation branches, we form two directional 
correlation matrices:
\begin{equation}
\begin{aligned}
C^{\mathrm{gen}\rightarrow\mathrm{ref}}_t
&= \mathrm{softmax}\!\left(
\frac{
\bar{Q}^{\mathrm{gen}}_{t-1}
(\bar{K}^{\mathrm{ref}}_{t-1})^\top
}{\sqrt{d}}
\right), \\
C^{\mathrm{ref}\rightarrow\mathrm{gen}}_t
&= \mathrm{softmax}\!\left(
\frac{
\bar{Q}^{\mathrm{ref}}_{t-1}
(\bar{K}^{\mathrm{gen}}_{t-1})^\top
}{\sqrt{d}}
\right),
\end{aligned}
\end{equation}

The subscripts $\mathrm{gen}$ and $\mathrm{ref}$ correspond to features from
the generation branch and the reference-inverted branch, respectively. The arrow notation indicates the direction in which correspondences are computed.
To improve reliability and reduce asymmetric matching artifacts, we 
combine both directions into a symmetric correlation map:
\begin{equation}
\hat{C}_t
= \frac{1}{2}
\left(
C^{\mathrm{gen}\rightarrow\mathrm{ref}}_t
+
(C^{\mathrm{ref}\rightarrow\mathrm{gen}}_t)^\top
\right).
\label{eq:symmetric-correlation}
\end{equation}

From $\hat{C}_t$, the most correlated token pairs are identified to compute 
relative spatial displacements between branches, producing bidirectional flows 
$F^{\mathrm{gen}\rightarrow\mathrm{ref}}_t$ and 
$F^{\mathrm{ref}\rightarrow\mathrm{gen}}_t$.  
These flows capture dense semantic correspondences between the two branches, 
which serve as the basis for value warping and cycle-consistency masking.

\paragraph{Selective and Consistent Appearance Injection.}
To ensure appearance is injected only into reliable and semantically relevant regions, we use two complementary masks, a foreground mask $M_{\mathrm{fg}}$ 
and a cycle-consistency mask $M_{\mathrm{cc}}$.

The foreground mask isolates subject-relevant regions within the generation branch.
We exploit the attention interactions between the subject text token and video tokens accumulated across denoising steps.
In MM-DiT architectures, where text and video tokens share a unified self-attention space, cross-modal interactions appear in two directions, \textit{text-to-video} (T$\!\rightarrow$V) and \textit{video-to-text} (V$\!\rightarrow$T).
Prior work~\cite{kim2025seg4diffunveilingopenvocabularysegmentation, jin2025matrixmasktrackalignment} has shown that
the V$\!\rightarrow$T attention region captures more complete and spatially coherent object extents,
making it more suitable for identifying subject-aligned regions.
Accordingly, $M_{\mathrm{fg}}$ is constructed by averaging the V$\!\rightarrow$T attention map across denoising steps
and selecting tokens whose averaged activation exceeds a threshold $\tau_{\mathrm{fg}}$.

To eliminate remaining unreliable matches, 
we compute a cycle-consistency mask using forward and reverse flows.  
The cycle-consistency error and corresponding binary mask are defined as
\begin{equation}
\begin{aligned}
\mathcal{E}_{\mathrm{cc}}
&=
\left\|
\mathcal{W}\!\left(
F^{\mathrm{gen}\rightarrow\mathrm{ref}}_t;\,
F^{\mathrm{ref}\rightarrow\mathrm{gen}}_t
\right)
\right\|, \\
M_{\mathrm{cc}}
&=
\!\left(
\mathcal{E}_{\mathrm{cc}} < \tau_{\mathrm{cc}} \cdot H \cdot 
\frac{\sum M_{\mathrm{fg}}}{H \times W}
\right),
\end{aligned}
\label{eq:cycle-mask}
\end{equation}
where $\mathcal{W}(\cdot;\cdot)$ denotes warping using the given flow pair. $\tau_{\mathrm{cc}}$ is a fixed threshold, scaled by the proportion of the
foreground region to adaptively determine the final threshold value.
Tokens with small $\mathcal{E}_{\mathrm{cc}}$ values are regarded as cycle-consistent.

The final injection region is defined as the intersection
\begin{equation}
M_t = M_{\mathrm{fg}} \odot M_{\mathrm{cc}}.
\end{equation}
This region restricts appearance injection to subject-relevant areas and prevents leakage into the background.
It also ensures that features are injected only where correspondences are reliable, avoiding erroneous or unstable transfer.

\paragraph{Value Feature Warping.}
Prior work~\cite{nam2024dreammatcher} has shown that 
the \textit{value} features in attention layers primarily encode appearance information.
We exploit this property by transferring fine appearance details 
from the reference branch to the generation branch through semantic flow-based warping:
\begin{equation}
V^{\mathrm{warp}}_{t, \mathrm{vid}}
=
\mathcal{W}\!\left(
V_{t,\mathrm{vid}}^{\mathrm{ref}};\,
F_{t}^{\mathrm{ref}\rightarrow\mathrm{gen}}
\right),
\end{equation}
where $\mathcal{W}(\cdot;\cdot)$ denotes warping guided by the flow 
$F_{t}^{\mathrm{ref}\rightarrow\mathrm{gen}}$.
In this formulation, $V^{\mathrm{ref}}_{t,\mathrm{vid}}$ and 
$V^{\mathrm{gen}}_{t,\mathrm{vid}}$ denote the video-token value features of the
reference and generation branches at denoising step $t$, respectively.

To inject appearance only into reliable subject regions, 
the warped values are blended with the generation branch values 
using the injection mask $M_t$:
\begin{equation}
\hat{V}^{\mathrm{gen}}_{\mathrm{vid}}
=
M_t
\odot
V^{\mathrm{warp}}_{t, \mathrm{vid}}
+
(1-M_t)
\odot
V^{\mathrm{gen}}_{t, \mathrm{vid}}.
\end{equation}

Finally, the updated video-token values are concatenated 
with the original text-token values and fed into the Multi-Modal Attention (MMA) layer:
\begin{equation}
\mathrm{MMA}([X;C_T])
=
\mathrm{softmax}\!\left(
\frac{QK^{\top}}{\sqrt{d}}
\right)
[\hat{V}^{\mathrm{gen}}_{\mathrm{vid}};V^{\mathrm{gen}}_{\mathrm{text}}].
\end{equation}
This process injects fine-grained appearance information, 
enhancing subject fidelity and improving overall identity preservation.

%% file: sec/4_experiment.tex
\begin{figure*}[t]
    \centering
    \includegraphics[width=\linewidth]{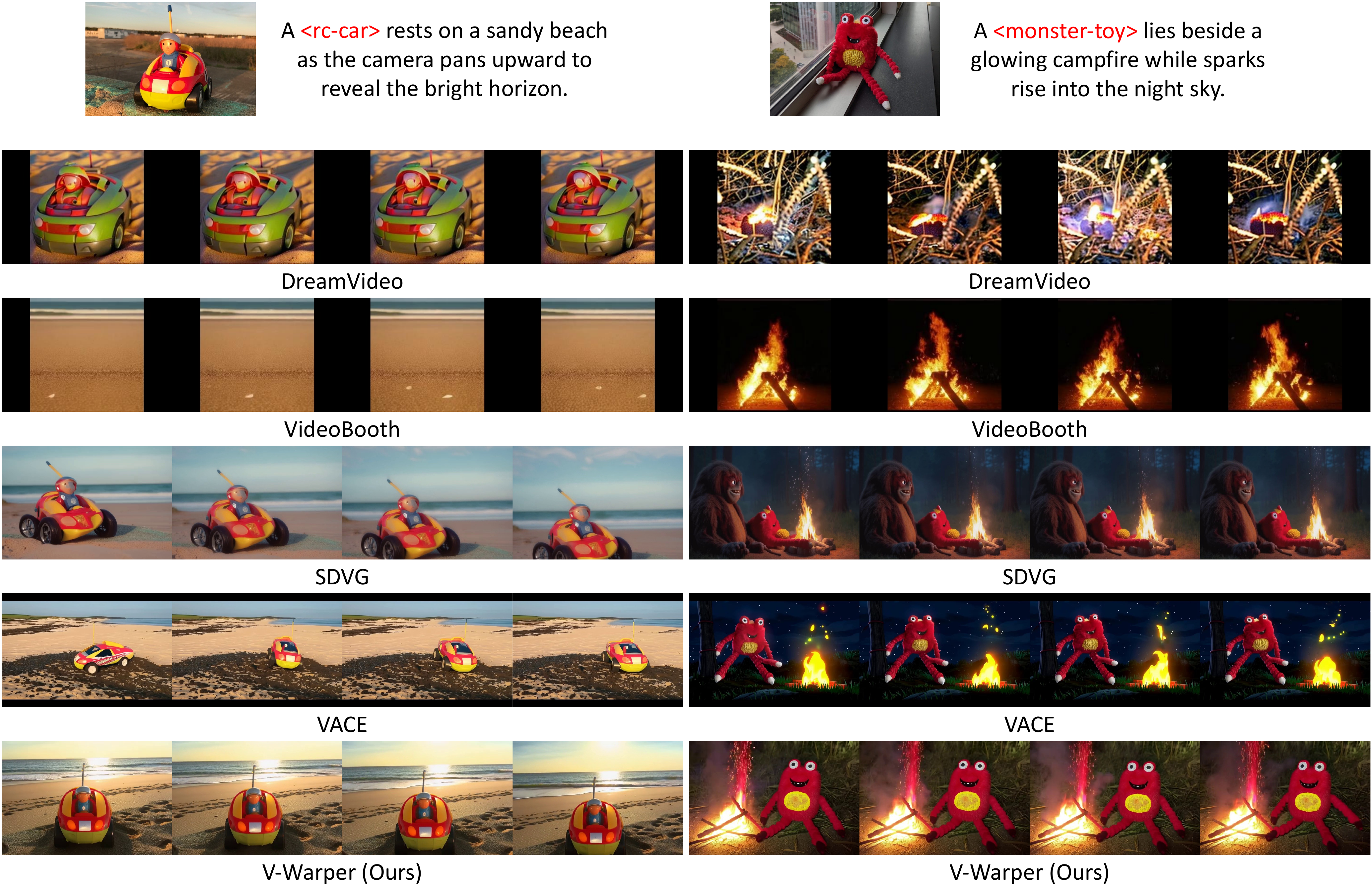}
    \caption{\textbf{Qualitative comparison with baselines.}
Across diverse subjects and prompts, \ours consistently preserves
fine appearance details such as color, texture, and geometry, whereas
prior video personalization methods often drift toward generic shapes,
lose distinctive identity cues, or introduce structural inconsistencies.
By combining coarse appearance adaptation with correspondence-guided
value warping, \ours maintains the closest match to the reference
subject while generating stable and visually coherent videos.}
    \label{fig:figure5_qualitative}
\end{figure*}

\section{Experiments}
\label{sec:experiments}

\subsection{Experimental Settings}
\paragraph{Datasets.}
We evaluate our method on a personalized video generation benchmark constructed from the DreamBooth dataset~\cite{ruiz2023dreambooth}. 
We select 10 subjects, each provided with 5 reference images. 
To assess robustness under different generative conditions, we design a prompt set that covers three dynamic scenarios, namely background transitions, subject motion, and camera motion.
For each subject, 10 prompts are generated using ChatGPT, resulting in a diverse and consistent evaluation suite.

\paragraph{Evaluation Metrics.}
Following prior video personalization works~\cite{wei2023dreamvideo, jiang2024videobooth, kim2025subject, huang2025videomage}, we use three embedding-based similarity metrics to evaluate identity preservation and text alignment.
CLIP-I measures global appearance fidelity and identity similarity between generated frames and reference images using CLIP~\cite{pmlr-v139-radford21a_clip}.
DINO-I evaluates fine-grained appearance fidelity using self-supervised features extracted with DINO~\cite{Caron_2021_ICCV_dino}.
CLIP-T measures text–video alignment by comparing frame embeddings with the prompt embeddings.

\paragraph{Comparison Methods.}
We compare \ours with two categories of video personalization approaches.
The first category comprises instance-specific optimization methods, which require per-subject fine-tuning to learn identity-specific parameters. DreamVideo~\cite{wei2023dreamvideo} represents this group, adapting identity through subject-specific optimization of an ID adapter.
The second category includes large-scale pretrained personalization models that leverage extensive video–text data to learn a generalizable identity prior.
VideoBooth~\cite{jiang2024videobooth}, Subject-Driven Video Generation~\cite{kim2025subject}, and VACE~\cite{jiang2025vace} fall into this group and serve as strong pretrained baselines.

\subsection{Experimental Results}
\paragraph{Quantitative Results}
\input{table/quantitative_result_w_baseline}
As shown in Table~\ref{tab:baseline_quantitative_results}, \ours achieves the highest scores on both CLIP-I and DINO-I, indicating the strongest preservation of subject appearance among all compared methods.
The improvement is consistent across diverse prompts and motion scenarios.
Such consistency shows that the coarse-to-fine design effectively captures both global identity and fine-grained texture details.
Our approach also achieves competitive CLIP-T performance, ranking second among all methods, indicating that improvements in appearance fidelity are attained without compromising text–video alignment.
Taken together, these results show that \ours delivers state-of-the-art identity fidelity while preserving prompt consistency, all without relying on heavy training or additional video data.





\paragraph{Qualitative Results}
Figure~\ref{fig:figure5_qualitative} compares \ours with representative
video–personalization baselines. Earlier optimization-based or fine-tuned
approaches such as DreamVideo and VideoBooth often fail to reproduce
identity-defining structures, leading to noticeable drift from the reference
appearance.

More advanced pretrained methods, including SDVG and VACE, exhibit improved
robustness but still show characteristic failure modes. 
Common issues include distorted geometry, as seen when SDVG deforms the body
shape of the RC car. Another issue is inconsistent texture patterns across
frames, where VACE loses the RC car’s side markings after the initial frames.
A further failure case is the generation of spurious parts, such as VACE
producing an extra leg in the monster-toy sequence. \ours delivers consistently more faithful identity reproduction.
The generated subjects maintain correct proportions, distinctive texture
patterns, and coherent appearance across frames. Our
correspondence-guided value injection transfers appearance only into
semantically matched regions, which preserves the model’s temporal priors
while enriching fine-grained visual details.

Overall, \ours produces temporally stable and identity-accurate
personalized videos, preserving fine-grained appearance cues that existing
baselines fail to maintain. Additional qualitative comparisons and video
samples are provided in Appendix~\ref{sec:add_results}.

\paragraph{User Study.}
We conducted a user study comparing \ours with the strongest competing baseline, VACE. 
Participants evaluated video pairs based on text alignment, subject fidelity, motion fidelity, 
and overall video quality. A total of 40 comparisons were rated by 20 participants. 
Additional details on the protocol and interface are provided in Appendix~\ref{sec:userstudy}. As shown in Fig.~\ref{fig:user_study}, \ours is consistently preferred over VACE across
all evaluation criteria, including text alignment, subject fidelity, motion fidelity, and overall
video quality. These results confirm that our method achieves stronger appearance preservation
and better perceptual quality.

\begin{figure}[t]
    \centering
    \includegraphics[width=\columnwidth]{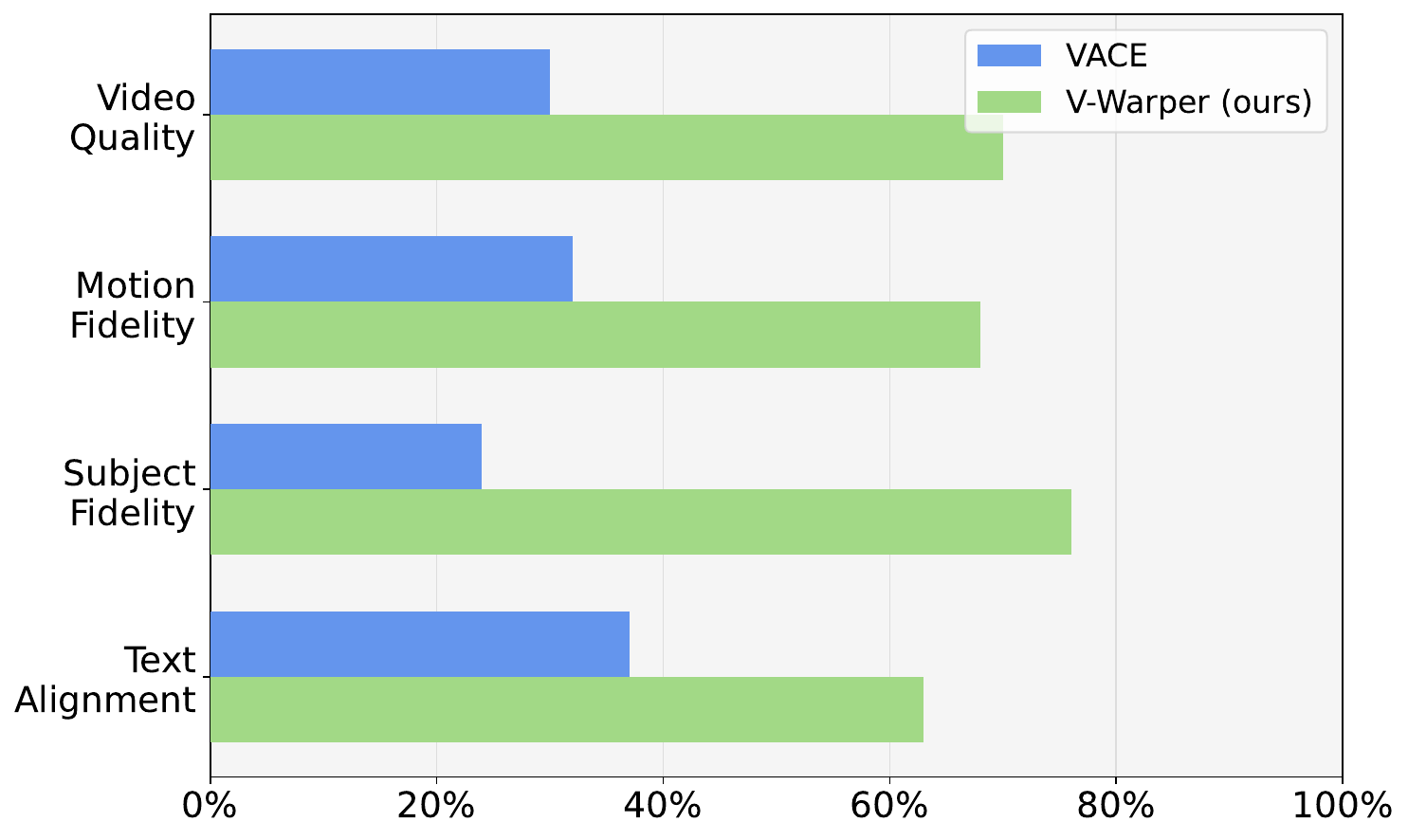}
    \caption{\textbf{User study results.}
    Participants consistently prefer \ours over VACE across all four 
    evaluation criteria—Text Alignment, Subject Fidelity, Motion Fidelity, 
    and Video Quality—highlighting its superior appearance preservation 
    and overall perceptual quality.}
    \label{fig:user_study}
\end{figure}

\input{table/quantitiave_result_abl}

\subsection{Ablation Studies}
We analyze the role of each component in our framework using the quantitative results in Table~\ref{tab:abl_quantitative_results} together with the qualitative
visualizations in Figure~\ref{fig:ablation_vis}.
(I) Coarse Appearance Adaptation,
(II) Value Warping,
and (III) Masked Warping (Full Model)
represent the three evaluation settings. (I) provides a reasonable identity initialization but lacks fine texture or color fidelity, as only coarse subject characteristics are learned from static images. (II) introduces value feature transfer, which significantly boosts identity metrics by propagating high-frequency appearance cues from the reference images.
However, as shown in Figure~\ref{fig:ablation_vis} (c), applying warping across all tokens also transfers unreliable matches. The clock develops an incorrect metallic connector on its top, and beach regions are overwritten with white-cloth textures from the reference image.
In our masked warping (III), Although identity scores slightly drop compared to (II) due to selective injection, qualitative fidelity improves over (I), and unwanted distortions are substantially reduced.
This results in better text–video alignment, evidenced by higher CLIP-T scores than (II).

Overall, these results highlight that value warping is crucial for injecting fine appearance details, while masking is essential for stable and clean personalization.
The full model (III) achieves the best overall balance between identity preservation and prompt consistency.

%% file: table/quantitative_result_w_baseline.tex
\begin{table}[t]
\centering
\small 
\setlength{\tabcolsep}{5pt} 
\begin{tabular}{lccc}
\toprule
Method & $I_{\text{DINO}} \uparrow$ & $I_{\text{CLIP}} \uparrow$ & $T_{\text{CLIP}} \uparrow$ \\
\midrule
DreamVideo~\cite{wei2023dreamvideo} & 0.322 & 0.641 & 0.290 \\
VideoBooth~\cite{jiang2024videobooth}      & 0.349 & 0.634 & 0.272 \\
SDVG~\cite{kim2025subject}            & \underline{0.661} & 0.787 & 0.294 \\
VACE~\cite{jiang2025vace}            & 0.651 & \underline{0.796} & \textbf{0.326} \\
\midrule
\textbf{\ours}                       & \textbf{0.738} & \textbf{0.825} & \underline{0.297} \\
\bottomrule
\end{tabular}
\caption{\textbf{Quantitative comparison with baselines.}
\ours shows the strongest identity preservation across both DINO-I and CLIP-I, surpassing all baselines by a significant margin. 
While VACE achieves a slightly higher CLIP-T score, \ours maintains competitive text alignment without relying on large-scale video finetuning, demonstrating the overall effectiveness of our coarse-to-fine personalization strategy.}\vspace{-5pt}
\label{tab:baseline_quantitative_results}
\end{table}

%% file: table/quantitiave_result_abl.tex
\begin{table}
\centering
\resizebox{\linewidth}{!}{%
\begin{tabular}{l l c c c}
\toprule
 & \textbf{Component} 
 & $\mathbf{I_{DINO}} \uparrow$
 & $\mathbf{I_{CLIP}} \uparrow$
 & $\mathbf{T_{CLIP}} \uparrow$ \\ 
\midrule
(I) & Coarse Appearance Adaptation & 0.645 & 0.791 & 0.320 \\
(II) & (I)+Value Warping               & 0.701 & 0.809 & 0.278 \\
(III) & (II)+Masking \textbf{(Ours)}             & 0.656 & 0.806 & 0.320 \\
\bottomrule
\end{tabular}%
}
\caption{\textbf{Component analysis}. 
Adding Value Warping (II) markedly improves identity similarity but reduces text alignment. 
Our Masking strategy (III) suppresses identity leakage and recovers text–image consistency, 
yielding the most balanced performance across all metrics.}
\label{tab:abl_quantitative_results}
\end{table}

%% file: sec/5_conclusion.tex
\begin{figure}[t]
    \centering
    \includegraphics[width=1.00\linewidth]{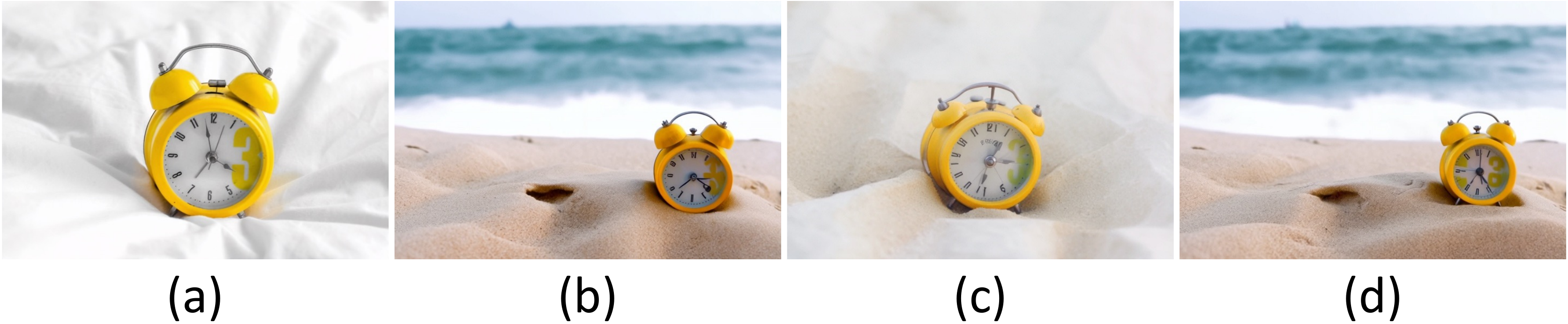}
    \caption{\textbf{Component analysis.}
    (a) Reference, (b) Coarse adaptation,  
    (c) + Value warping, (d) + Masking (full model).  
    Value warping enhances fine appearance details, while masking removes background leakage for the cleanest result.}
    \label{fig:ablation_vis}
\end{figure}

\vspace{-5pt}

\section{Conclusion}
\label{sec:conclusion}


We presented \ours, a training-free coarse-to-fine framework for subject-driven video personalization.
A lightweight image-based adaptation captures global subject identity in a coarse manner. 
Building on this foundation, an inference-time refinement stage injects fine-grained appearance using semantic correspondences and masked value warping.
This design preserves the model’s temporal prior and ensures that appearance is enhanced only in semantically reliable regions.
Extensive experiments show that \ours substantially improves appearance fidelity while maintaining strong prompt alignment and motion dynamics. These results demonstrate that \ours offers an efficient solution without large-scale video-based finetuning.

%% file: sec/X_suppl.tex
\clearpage
\renewcommand{\thesection}{\Alph{section}}
\renewcommand{\thefigure}{A.\arabic{figure}}
\renewcommand{\thetable}{A.\arabic{table}}

\setcounter{section}{0}
\setcounter{figure}{0}
\setcounter{table}{0}
\section*{\Large Appendix}

This appendix provides additional implementation details,
analysis, evaluations, and extended qualitative results that complement
the main paper. The content is organized as follows:

\begin{itemize}
    \item \textbf{Sec.~A}: Implementation details for coarse appearance
    adaptation and fine appearance injection.
    \item \textbf{Sec.~B}: Analysis of semantic matching behavior
    in diffusion transformers and additional ablation on alternative
    feature-injection strategies.
    \item \textbf{Sec.~C}: Additional evaluations, including full VBench
    results and a detailed description of the user study setup.
    \item \textbf{Sec.~D}: Extended qualitative results across all subjects
    and prompts.
    \item \textbf{Sec.~E}: Limitations and discussion.
\end{itemize}

\vspace{10pt}

\section{Implementation Details}
\label{sec:impl_details}

\paragraph{Generation Setup.}
We use CogVideoX-5B~\cite{yang2024cogvideox} with its official pretrained weights and do not modify the underlying architecture.
All videos are generated at 25 frames, with
50 denoising steps and 16 fps.
All experiments are performed on a single RTX~A6000 GPU.

\paragraph{Coarse Appearance Adaptation.}
During coarse adaptation, LoRA~\cite{hu2021lora} modules are attached to
the key, value, and output projections of every MM-DiT attention layer.
We use rank 128 and optimize LoRA parameters using AdamW with a learning
rate of $1\times 10^{-4}$. A subject-specific learnable token embedding is
trained jointly with a separate learning rate of $5\times 10^{-4}$.

\paragraph{Fine Appearance Injection.}
During inference, we inject high-frequency appearance details via value--feature warping.
We first construct a parallel reference branch by repeating the reference
image along the temporal axis to match the number of frames in the generation branch.
We then obtain its latent trajectory through DDIM~\cite{song2022_ddim} inversion so that both
branches share synchronized denoising steps.
Cross-branch token-level correspondences are estimated from Layer~12
RoPE-free query and key features, which provide the most reliable semantic
alignment among all evaluated representations.
Value warping is applied between denoising steps 3--19 and across MM-DiT layers 20--29. 
These ranges balance structural stability in early timesteps and sufficient refinement in later stages while targeting layers that best capture high-frequency appearance details.

Foreground masks are derived by aggregating subject text–token attention across denoising steps and layers. For each timestep~$t$, we average the softmax-normalized attention between the subject text token and all video tokens in the video–text attention blocks.  
Positions whose aggregated score exceeds $\tau_{\mathrm{fg}}=0.3$ are retained.
Cycle-consistency masks are computed from bidirectional flows by first
defining the cycle-consistency error as the deviation between each token’s
original position and its round-trip reconstructed coordinate. Tokens with cycle-consistency error below $\tau_{\mathrm{cc}} = 0.1$ scaled by the
foreground ratio are preserved, ensuring that only geometrically coherent
correspondences participate in the final value injection.

\section{Analysis}
\label{sec:analysis}

\subsection{Matching Performance}
\label{sec:matching}

Here we describe the protocol used to evaluate semantic correspondence quality. Following the same evaluation benchmark described in the main paper, we use 10 subjects from the DreamBooth~\cite{ruiz2023dreambooth} dataset, each paired with 10 prompts, yielding 100 evaluation cases in total. For each subject and prompt, videos are generated using our dual-branch inference setup, and dense image correspondences are obtained by applying DINOv3~\cite{simeoni2025dinov3} between the generated video frames and the reference image.  
These matches serve as pseudo ground truth for evaluating the model’s internal correspondence predictions.  
At denoising steps $t\in\{10,20,30,40\}$, we extract cross-branch flows from all MM-DiT layers using three feature types, specifically intermediate activations, raw query–key pairs, and RoPE-free query–key pairs.
All evaluations are restricted to subject-foreground tokens, obtained via Grounded-SAM segmentation.

Correspondence accuracy is measured using PCK~\cite{yang2012articulated} with a threshold 
$\alpha=0.05$ relative to image resolution.  
The resulting layer-wise PCK curves confirm that RoPE-free query–key features, particularly those extracted from Layer 12, provide the most reliable semantic correspondences among all evaluated representations.
These findings validate our design choice for the matching representation used in \ours's fine appearance injection stage.

\input{table/suppl_vbench}

\subsection{Appearance Injection Ablation}
\label{sec:appearance_ablation}

\begin{figure*}[t]
    \centering
    \includegraphics[width=0.85\textwidth]{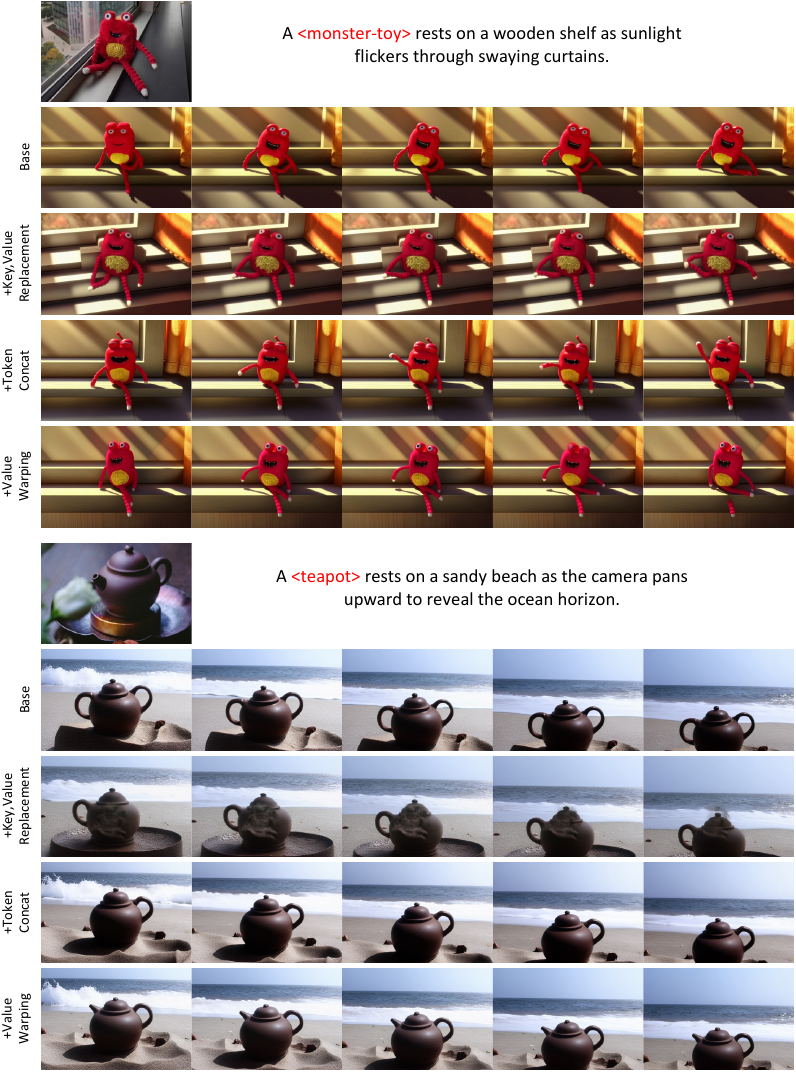}
    \caption{\textbf{Ablation of appearance injection strategies.}
    We compare Key--Value Replacement, Token Concatenation, and our
    Value Warping on identical coarse-adapted models.  
    Key--Value Replacement modifies attention routing and often produces
    pose distortion or artifacts.  
    Token Concatenation provides only weak implicit conditioning and
    fails to reliably improve subject identity.  
    Our Value Warping selectively transfers appearance along
    semantically matched regions, achieving the most faithful and
    stable appearance preservation.}
    \label{fig:appearance_injection_ablation}
\end{figure*}

Beyond our value-feature warping mechanism, prior works on attention manipulation
have introduced several alternative appearance-injection strategies
\cite{shi2023masactrl,Cai_2025_CVPR,tan2025ominicontrol}. 
We compare three representative approaches built on the same
coarse-adapted CogVideoX-5B. 
The methods evaluated are 
(i) \emph{Key and Value Replacement}, 
(ii) \emph{Token Concatenation}, 
and 
(iii) our \emph{Value Warping}.  
Qualitative comparisons are presented in Fig.~\ref{fig:appearance_injection_ablation}.

\paragraph{Key--Value Replacement.}
A common strategy in attention-control methods
\cite{shi2023masactrl,Cai_2025_CVPR}
is to run foreground/background attentions separately and blend them.
We adapt this idea to our dual-branch setup.

Let $M^{\mathrm{ref}}$ denote the foreground mask on the reference
branch (subject regions) and $M^{\mathrm{gen}}$ the injection mask on the
generation branch.
Before performing MMA, we construct multimodal keys and values for each
branch by concatenating text- and video-token components:
\begin{equation}
\begin{aligned}
K^{\mathrm{gen}} &= 
\bigl[\,K^{\mathrm{gen}}_{\mathrm{text}} \,;\,
       K^{\mathrm{gen}}_{\mathrm{vid}} \,\bigr], \\
K^{\mathrm{ref}} &=
\bigl[\,K^{\mathrm{gen}}_{\mathrm{text}} \,;\,
       K^{\mathrm{ref}}_{\mathrm{vid}} \,\bigr], \\
V^{\mathrm{gen}} &= 
\bigl[\,V^{\mathrm{gen}}_{\mathrm{text}} \,;\,
       V^{\mathrm{gen}}_{\mathrm{vid}} \,\bigr], \\
V^{\mathrm{ref}} &=
\bigl[\,V^{\mathrm{gen}}_{\mathrm{text}} \,;\,
       V^{\mathrm{ref}}_{\mathrm{vid}} \,\bigr].
\end{aligned}
\end{equation}
Since text tokens have no appearance to transfer, their keys and values
always come from the generation branch.

Using queries from the generation branch, we compute two attentions:
a reference-conditioned foreground output and a generation-only background
output:
\begin{equation}
\begin{aligned}
f^{\mathrm{ref}} &= 
\mathrm{MMA}\!\left(
    Q^{\mathrm{gen}},
    K^{\mathrm{ref}},
    V^{\mathrm{ref}};\,
    M^{\mathrm{ref}}
\right), \\[4pt]
f^{\mathrm{gen}} &= 
\mathrm{MMA}\!\left(
    Q^{\mathrm{gen}},
    K^{\mathrm{gen}},
    V^{\mathrm{gen}};\,
    1 - M^{\mathrm{gen}}
\right).
\end{aligned}
\end{equation}

Finally, the blended output is produced only for the video--video
attention regions:
\begin{equation}
\bar{f}
= f^{\mathrm{ref}} \odot M^{\mathrm{gen}}
 + f^{\mathrm{gen}} \odot (1 - M^{\mathrm{gen}}).
\end{equation}

While this approach injects strong subject appearance, replacing
reference keys inside the foreground region alters the key distribution
that governs attention routing. This frequently disrupts the spatial
and temporal priors of the base model, leading to structural and motion
instability. As illustrated in Fig.~\ref{fig:appearance_injection_ablation},
Key--Value Replacement often forces the generation branch to imitate the
reference pose rather than follow the intended video dynamics, as seen
in the \textit{monster-toy} example. It can also introduce noticeable
geometric distortions, such as corrupted contours in the \textit{teapot}
generation. These behaviors indicate that the model’s internal geometry
and temporal consistency are compromised.

\paragraph{Token Concatenation.}
Another line of work concatenates reference-image tokens with
generation tokens and performs a single
attention pass over the joint sequence~\cite{tan2025ominicontrol}.
In training-based settings, spatially non-aligned tasks are handled using shifted
positional encoding, but applying such shifts in a training-free setup
pushes the model outside its learned RoPE distribution.
We therefore keep the original positions for reference tokens, leading
to weak and unreliable conditioning.

As shown in Fig.~\ref{fig:appearance_injection_ablation}, this implicit
conditioning seldom improves subject fidelity and may even degrade it.
Because no explicit correspondence guides where reference appearance
should be injected, the model incorrectly propagates appearance cues.
For instance, the \textit{monster-toy} exhibits distorted facial details
such as altered eye shape, and the \textit{teapot} still retains the
incorrect double-handle structure present in the base output.
These observations indicate that simple token concatenation is
insufficient for precise appearance transfer without an explicit
matching mechanism.

\paragraph{Value Warping.}
Instead of modifying keys or concatenating tokens, we preserve the
generation branch’s native attention routing and transfer appearance
through the warped value features alone.
Semantic correspondences, estimated as described in the main paper,
guide the injection of reference appearance so that updates occur only
within subject regions that exhibit reliable matches.

As illustrated in Fig.~\ref{fig:appearance_injection_ablation}, this
selective value transfer reliably enhances fine appearance details while
avoiding structural distortion. For example, the \textit{monster-toy}
case recovers the correct fluffy texture and eye shape that other
methods fail to preserve, and the \textit{teapot} example correctly
removes the erroneous double handles and restores the proper spout
structure. These improvements demonstrate that value warping provides
precise, correspondence-aware appearance injection, achieving faithful
and artifact-free identity preservation.

\section{Evaluation}
\label{sec:evaluation}

\subsection{VBench Evaluation}
\label{sec:vbench}

We report full VBench~\cite{huang2024vbench} results for \ours and
baseline models. \ours outperforms the optimization-based baseline
DreamVideo in all metrics except Background Consistency, and remains
competitive with large training-based systems despite requiring no
video-scale fine-tuning. Full scores are shown in
Table~\ref{tab:baseline_vbench}.

\subsection{User Study Details}
\label{sec:userstudy}

\begin{figure}[t]
    \centering
    \includegraphics[width=1.0\linewidth]{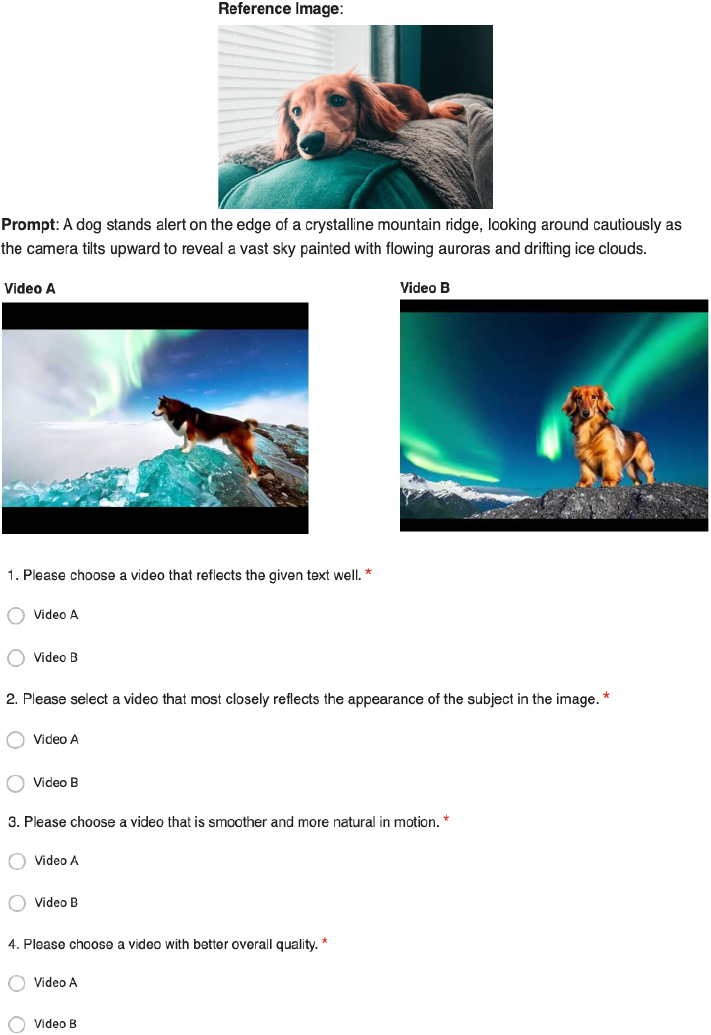}
    \caption{\textbf{User Study Interface.} Example evaluation screen showing the reference image, text prompt, and two generated videos presented in randomized order for pairwise preference selection.}
    \label{fig:user_study_interface}
\end{figure}

An example evaluation screen is shown in Figure~\ref{fig:user_study_interface}. We conducted a paired human-preference study comparing \ours with VACE to assess subject fidelity, text alignment, motion fidelity, and overall video quality. For each question, participants were shown a reference image, the text prompt, and two generated videos in randomized order, and were asked to choose the preferred result for each criterion. A total of 20 participants responded to 40 comparison questions each, yielding 800 pairwise evaluations. Video order and question order were fully randomized to avoid bias. As summarized in the main paper, participants consistently favored \ours, especially in subject fidelity, confirming its stronger appearance preservation in real perceptual judgments.
These interface settings correspond to the user study outcomes summarized
in the main paper, where \ours is consistently preferred across all
evaluation criteria.

\section{Additional Results}
\label{sec:add_results}

\begin{figure*}[t]
    \centering
    \includegraphics[width=0.73\textwidth]{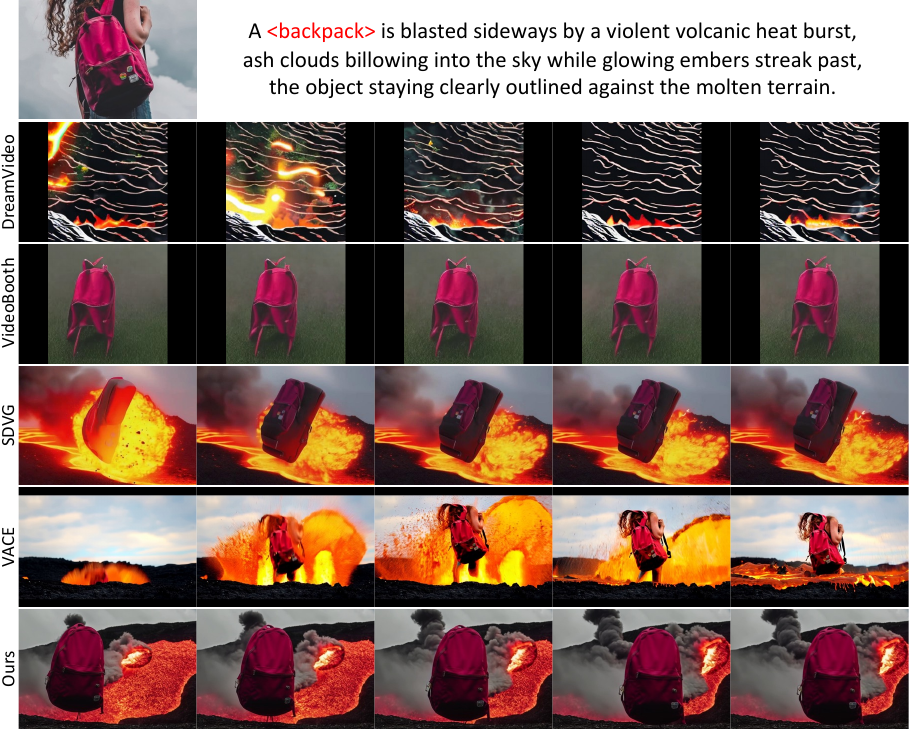}
    \caption{Additional qualitative results (Set~1).}
    \label{fig:add_qual_1}
\end{figure*}

\begin{figure*}[t]
    \centering
    \includegraphics[width=0.73\textwidth]{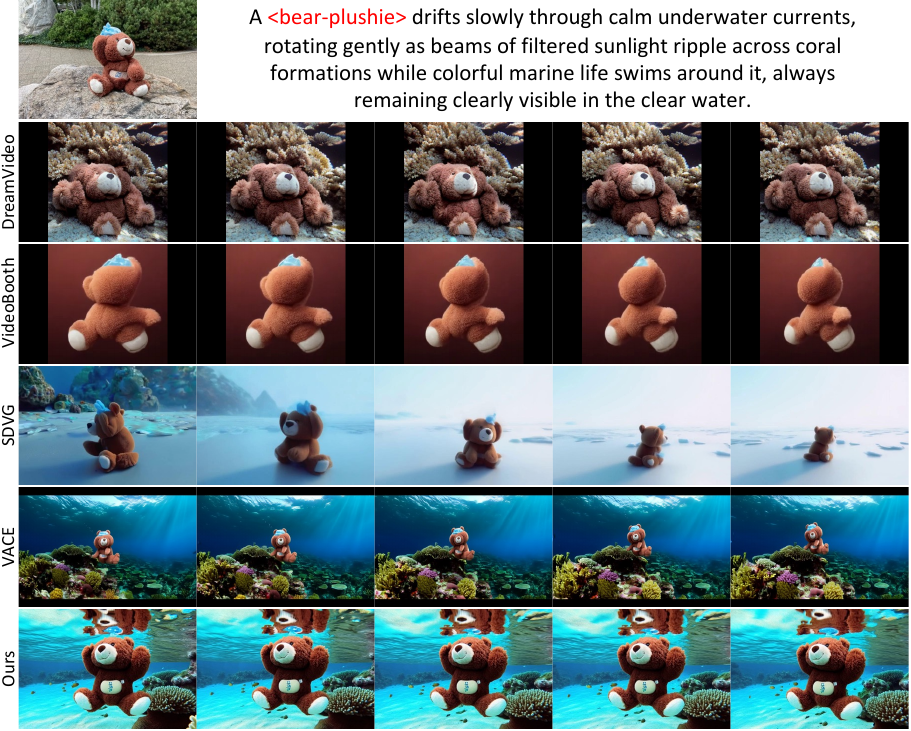}
    \caption{Additional qualitative results (Set~2).}
    \label{fig:add_qual_2}
\end{figure*}

\begin{figure*}[t]
    \centering
    \includegraphics[width=0.73\textwidth]{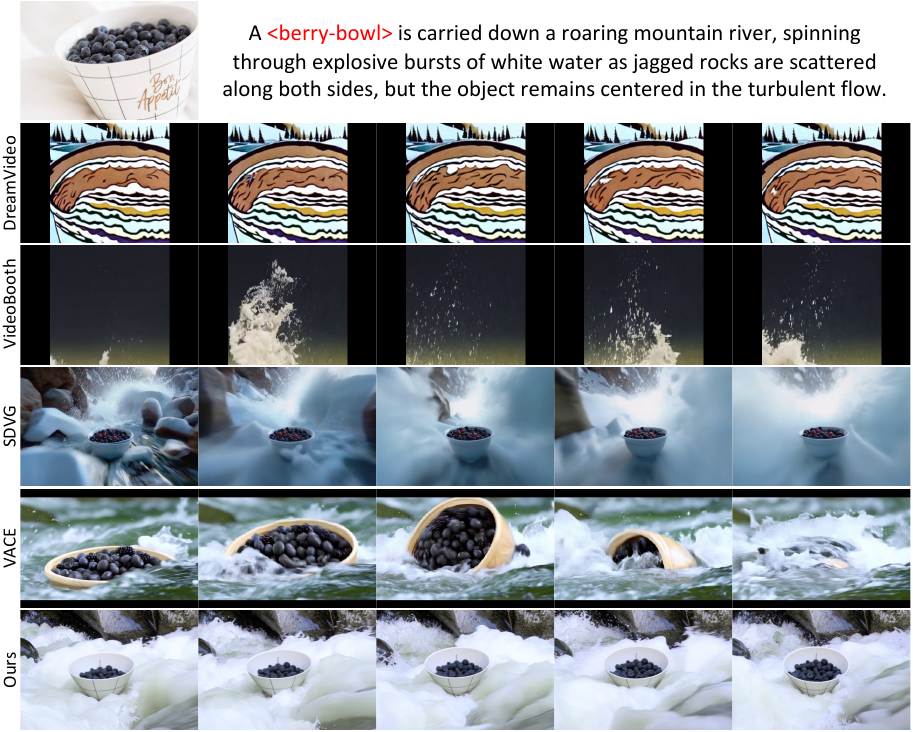}
    \caption{Additional qualitative results (Set~3).}
    \label{fig:add_qual_3}
\end{figure*}

\begin{figure*}[t]
    \centering
    \includegraphics[width=0.73\textwidth]{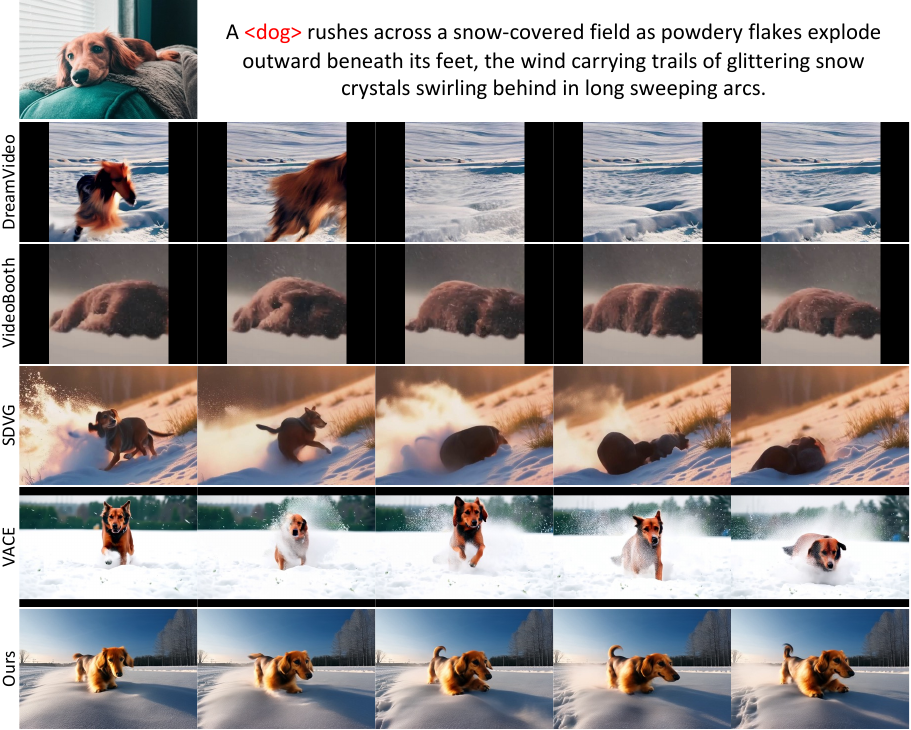}
    \caption{Additional qualitative results (Set~4).}
    \label{fig:add_qual_4}
\end{figure*}

Figures~\ref{fig:add_qual_1}--\ref{fig:add_qual_4}
provide extended qualitative examples across diverse subjects,
prompt categories, and motion scenarios.
\ours consistently preserves fine-grained identity cues and
generates stable motion under challenging settings.

\section{Limitations}
\label{sec:limit}

\ours demonstrates strong appearance preservation and competitive
video quality, but several limitations remain:

\paragraph{Per-subject optimization.}
Our method performs a lightweight subject-specific optimization during the coarse adaptation stage.
Since this step uses only a few reference images and does not rely on any subject-specific video data, it is substantially more efficient than video-based fine-tuning approaches.
However, the requirement of per-subject optimization remains a limitation, and removing this dependency represents an important direction for future work.

\paragraph{Motion expressiveness.}
Although our design preserves most of the base model’s motion prior,
some loss of motion richness may still occur. This challenge is common
in video personalization and may be further mitigated with improved
masking or hybrid temporal modeling.

Despite these limitations, \ours provides an effective and efficient
solution for high-fidelity video personalization without large-scale
video training.


%% file: table/suppl_vbench.tex

\begin{table*}[t]
\centering
\resizebox{\textwidth}{!}{%
\begin{tabular}{lccccccc}
\toprule
Method &
Subject Cons. $\uparrow$ &
Background Cons. $\uparrow$ &
Motion Smooth. $\uparrow$ &
Dynamic Degree $\uparrow$ &
Aesthetic Quality $\uparrow$ &
Imaging Quality $\uparrow$ &
Temporal Flicker $\uparrow$ \\
\midrule
\multicolumn{8}{l}{\textit{Training-based video personalization}} \\
VideoBooth      & 0.9143 & 0.9494 & 0.9649 & 0.6900 & 0.4483 & 56.0102 & 0.9550 \\
SVGD            & 0.9811 & 0.9801 & 0.9922 & 0.2100 & 0.6498 & 64.9768 & 0.9889 \\
VACE            & 0.9685 & 0.9751 & 0.9829 & 0.4500 & 0.6749 & 67.8530 & 0.9650 \\
\midrule
\multicolumn{8}{l}{\textit{Optimization-based video personalization}} \\
DreamVideo      
& 0.9591 
& \textbf{0.9766} 
& 0.9734 
& 0.1400 
& 0.5111 
& 62.4017 
& 0.9630 \\
\midrule
\textbf{\ours (Ours)}
& \textbf{0.9866} 
& 0.9750 
& \textbf{0.9866} 
& \textbf{0.5100} 
& \textbf{0.6074} 
& \textbf{70.6617} 
& \textbf{0.9774} \\
\bottomrule
\end{tabular}%
}
\caption{
\textbf{VBench comparison.}
\ours achieves state-of-the-art performance among optimization-based methods, 
outperforming DreamVideo in all metrics except Background Consistency.  
Compared with training-based approaches, \ours attains the best scores in 
Subject Consistency and Imaging Quality, 
while remaining competitive across all other temporal and perceptual metrics—despite requiring no large-scale video training.
}
\label{tab:baseline_vbench}
\end{table*}

%% file: main.bib
@article{ruiz2023dreambooth,
  title={DreamBooth: Fine Tuning Text-to-Image Diffusion Models for Subject-Driven Generation},
  author={Ruiz, Nataniel and Li, Yuanzhen and Jampani, Varun and Pritch, Yael and Rubinstein, Michael and Aberman, Kfir},
  journal={arXiv preprint arXiv:2208.12242},
  year={2023}
}

@article{gal2022textual,
  title={An Image is Worth One Word: Personalizing Text-to-Image Generation using Textual Inversion},
  author={Gal, Rinon and Alaluf, Yuval and Atzmon, Yuval and Patashnik, Or and Bermano, Amit H and Chechik, Gal and Cohen-Or, Daniel},
  journal={arXiv preprint arXiv:2208.01618},
  year={2022}
}

@article{yang2024cogvideox,
  title={CogVideoX: Text-to-Video Diffusion Models with Transformers},
  author={Yang, Fan and Zhou, Xingyi and Shen, Zhi and et al.},
  journal={arXiv preprint arXiv:2408.06072},
  year={2024}
}

@article{kong2024hunyuanvideo,
  title={HunyuanVideo: A Systematic Framework for Large Video Diffusion Models},
  author={Kong, Ming and Zhang, Xiaohan and et al.},
  journal={arXiv preprint arXiv:2412.03670},
  year={2024}
}

@article{wei2023dreamvideo,
  title={DreamVideo: Composing Your Dream Videos with Customized Subject and Motion},
  author={Wei, Dongxu and et al.},
  journal={arXiv preprint arXiv:2312.04433},
  year={2023}
}

@article{jiang2025vace,
  title={VACE: A Unified Framework for Video Creation and Editing with Diffusion Models},
  author={Jiang, Yu and et al.},
  journal={arXiv preprint arXiv:2503.07598},
  year={2025}
}

@inproceedings{huang2025videomage,
  title={Videomage: Multi-subject and motion customization of text-to-video diffusion models},
  author={Huang, Chi-Pin and Wu, Yen-Siang and Chung, Hung-Kai and Chang, Kai-Po and Yang, Fu-En and Wang, Yu-Chiang Frank},
  booktitle={Proceedings of the Computer Vision and Pattern Recognition Conference},
  pages={17603--17612},
  year={2025}
}

@inproceedings{jiang2024videobooth,
  title={Videobooth: Diffusion-based video generation with image prompts},
  author={Jiang, Yuming and Wu, Tianxing and Yang, Shuai and Si, Chenyang and Lin, Dahua and Qiao, Yu and Loy, Chen Change and Liu, Ziwei},
  booktitle={Proceedings of the IEEE/CVF Conference on Computer Vision and Pattern Recognition},
  pages={6689--6700},
  year={2024}
}

@article{ye2023ip,
  title={Ip-adapter: Text compatible image prompt adapter for text-to-image diffusion models},
  author={Ye, Hu and Zhang, Jun and Liu, Sibo and Han, Xiao and Yang, Wei},
  journal={arXiv preprint arXiv:2308.06721},
  year={2023}
}

@inproceedings{nam2024dreammatcher,
  title={Dreammatcher: Appearance matching self-attention for semantically-consistent text-to-image personalization},
  author={Nam, Jisu and Kim, Heesu and Lee, DongJae and Jin, Siyoon and Kim, Seungryong and Chang, Seunggyu},
  booktitle={Proceedings of the IEEE/CVF Conference on Computer Vision and Pattern Recognition},
  pages={8100--8110},
  year={2024}
}

@inproceedings{peebles2023scalable,
  title={Scalable diffusion models with transformers},
  author={Peebles, William and Xie, Saining},
  booktitle={Proceedings of the IEEE/CVF international conference on computer vision},
  pages={4195--4205},
  year={2023}
}

@inproceedings{shin2025large,
  title={Large-scale text-to-image model with inpainting is a zero-shot subject-driven image generator},
  author={Shin, Chaehun and Choi, Jooyoung and Kim, Heeseung and Yoon, Sungroh},
  booktitle={Proceedings of the Computer Vision and Pattern Recognition Conference},
  pages={7986--7996},
  year={2025}
}

@inproceedings{tan2025ominicontrol,
  title={Ominicontrol: Minimal and universal control for diffusion transformer},
  author={Tan, Zhenxiong and Liu, Songhua and Yang, Xingyi and Xue, Qiaochu and Wang, Xinchao},
  booktitle={Proceedings of the IEEE/CVF International Conference on Computer Vision},
  pages={14940--14950},
  year={2025}
}

@inproceedings{shi2023masactrl,
    author    = {Cao, Mingdeng and Wang, Xintao and Qi, Zhongang and Shan, Ying and Qie, Xiaohu and Zheng, Yinqiang},
    title     = {MasaCtrl: Tuning-Free Mutual Self-Attention Control for Consistent Image Synthesis and Editing},
    booktitle = {Proceedings of the IEEE/CVF International Conference on Computer Vision (ICCV)},
    month     = {October},
    year      = {2023},
    pages     = {22560-22570}
}

@inproceedings{xing2025motioncanvas,
  title={Motioncanvas: Cinematic shot design with controllable image-to-video generation},
  author={Xing, Jinbo and Mai, Long and Ham, Cusuh and Huang, Jiahui and Mahapatra, Aniruddha and Fu, Chi-Wing and Wong, Tien-Tsin and Liu, Feng},
  booktitle={Proceedings of the Special Interest Group on Computer Graphics and Interactive Techniques Conference Conference Papers},
  pages={1--11},
  year={2025}
}

@inproceedings{lei2025animateanything,
  title={Animateanything: Consistent and controllable animation for video generation},
  author={Lei, Guojun and Wang, Chi and Zhang, Rong and Wang, Yikai and Li, Hong and Xu, Weiwei},
  booktitle={Proceedings of the Computer Vision and Pattern Recognition Conference},
  pages={27946--27956},
  year={2025}
}

@article{feng2025personalize,
  title={Personalize anything for free with diffusion transformer},
  author={Feng, Haoran and Huang, Zehuan and Li, Lin and Lv, Hairong and Sheng, Lu},
  journal={arXiv preprint arXiv:2503.12590},
  year={2025}
}

@article{hu2025hunyuancustom,
  title={Hunyuancustom: A multimodal-driven architecture for customized video generation},
  author={Hu, Teng and Yu, Zhentao and Zhou, Zhengguang and Liang, Sen and Zhou, Yuan and Lin, Qin and Lu, Qinglin},
  journal={arXiv preprint arXiv:2505.04512},
  year={2025}
}

@article{kim2025subject,
  title={Subject-driven Video Generation via Disentangled Identity and Motion},
  author={Kim, Daneul and Zhang, Jingxu and Jin, Wonjoon and Cho, Sunghyun and Dai, Qi and Park, Jaesik and Luo, Chong},
  journal={arXiv preprint arXiv:2504.17816},
  year={2025}
}

@article{su2024roformer,
  title={Roformer: Enhanced transformer with rotary position embedding},
  author={Su, Jianlin and Ahmed, Murtadha and Lu, Yu and Pan, Shengfeng and Bo, Wen and Liu, Yunfeng},
  journal={Neurocomputing},
  volume={568},
  pages={127063},
  year={2024},
  publisher={Elsevier}
}

@article{nam2025emergent,
  title={Emergent Temporal Correspondences from Video Diffusion Transformers},
  author={Nam, Jisu and Son, Soowon and Chung, Dahyun and Kim, Jiyoung and Jin, Siyoon and Hur, Junhwa and Kim, Seungryong},
  journal={arXiv preprint arXiv:2506.17220},
  year={2025}
}

@article{blattmann2023stablevd,
  title={Stable Video Diffusion: Scaling Latent Video Diffusion Models to Large Datasets},
  author={A. Blattmann and Tim Dockhorn and Sumith Kulal and Daniel Mendelevitch and Maciej Kilian and Dominik Lorenz},
  journal={ArXiv},
  year={2023},
  volume={abs/2311.15127},
  url={https://api.semanticscholar.org/CorpusID:265312551}
}

@article{yang2024amg,
  title={AMG: Avatar Motion Guided Video Generation},
  author={Yang, Zhangsihao and Shan, Mengyi and Farazi, Mohammad and Zhu, Wenhui and Chen, Yanxi and Dong, Xuanzhao and Wang, Yalin},
  journal={arXiv preprint arXiv:2409.01502},
  year={2024}
}

@article{yu2025llia,
  title={LLIA--Enabling Low-Latency Interactive Avatars: Real-Time Audio-Driven Portrait Video Generation with Diffusion Models},
  author={Yu, Haojie and Wang, Zhaonian and Pan, Yihan and Cheng, Meng and Yang, Hao and Wang, Chao and Xie, Tao and Xu, Xiaoming and Wei, Xiaoming and Cai, Xunliang},
  journal={arXiv preprint arXiv:2506.05806},
  year={2025}
}

@article{wan2025wan,
  title={Wan: Open and advanced large-scale video generative models},
  author={Wan, Team and Wang, Ang and Ai, Baole and Wen, Bin and Mao, Chaojie and Xie, Chen-Wei and Chen, Di and Yu, Feiwu and Zhao, Haiming and Yang, Jianxiao and others},
  journal={arXiv preprint arXiv:2503.20314},
  year={2025}
}

@article{chen2025goku,
  title={Goku: Flow Based Video Generative Foundation Models.(2025)},
  author={Chen, Shoufa and Ge, Chongjian and Zhang, Yuqi and Zhang, Yida and Zhu, Fengda and Yang, Hao and Hao, Hongxiang and Wu, Hui and Lai, Zhichao and Hu, Yifei and others},
  journal={URL https://arxiv. org/abs/2502.04896},
  year={2025}
}

@article{hacohen2024ltx,
  title={Ltx-video: Realtime video latent diffusion},
  author={HaCohen, Yoav and Chiprut, Nisan and Brazowski, Benny and Shalem, Daniel and Moshe, Dudu and Richardson, Eitan and Levin, Eran and Shiran, Guy and Zabari, Nir and Gordon, Ori and others},
  journal={arXiv preprint arXiv:2501.00103},
  year={2024}
}

@article{hu2021lora,
  title={Lora: Low-rank adaptation of large language models. arXiv 2021},
  author={Hu, Edward J and Shen, Yelong and Wallis, Phillip and Allen-Zhu, Zeyuan and Li, Yuanzhi and Wang, Shean and Wang, Lu and Chen, Weizhu},
  journal={arXiv preprint arXiv:2106.09685},
  volume={10},
  year={2021}
}

@article{atzmon2024motion,
  title={Motion by Queries: Identity-Motion Trade-offs in Text-to-Video Generation},
  author={Atzmon, Yuval and Gal, Rinon and Tewel, Yoad and Kasten, Yoni and Chechik, Gal},
  journal={arXiv preprint arXiv:2412.07750},
  year={2024}
}

@article{simeoni2025dinov3,
  title={Dinov3},
  author={Sim{\'e}oni, Oriane and Vo, Huy V and Seitzer, Maximilian and Baldassarre, Federico and Oquab, Maxime and Jose, Cijo and Khalidov, Vasil and Szafraniec, Marc and Yi, Seungeun and Ramamonjisoa, Micha{\"e}l and others},
  journal={arXiv preprint arXiv:2508.10104},
  year={2025}
}

@article{yang2012articulated,
  title={Articulated human detection with flexible mixtures of parts},
  author={Yang, Yi and Ramanan, Deva},
  journal={IEEE transactions on pattern analysis and machine intelligence},
  volume={35},
  number={12},
  pages={2878--2890},
  year={2012},
  publisher={IEEE}
}

@inproceedings{huang2024vbench,
  title={Vbench: Comprehensive benchmark suite for video generative models},
  author={Huang, Ziqi and He, Yinan and Yu, Jiashuo and Zhang, Fan and Si, Chenyang and Jiang, Yuming and Zhang, Yuanhan and Wu, Tianxing and Jin, Qingyang and Chanpaisit, Nattapol and others},
  booktitle={Proceedings of the IEEE/CVF Conference on Computer Vision and Pattern Recognition},
  pages={21807--21818},
  year={2024}
}

@InProceedings{Cai_2025_CVPR,
    author    = {Cai, Minghong and Cun, Xiaodong and Li, Xiaoyu and Liu, Wenze and Zhang, Zhaoyang and Zhang, Yong and Shan, Ying and Yue, Xiangyu},
    title     = {DiTCtrl: Exploring Attention Control in Multi-Modal Diffusion Transformer for Tuning-Free Multi-Prompt Longer Video Generation},
    booktitle = {Proceedings of the IEEE/CVF Conference on Computer Vision and Pattern Recognition (CVPR)},
    month     = {June},
    year      = {2025},
    pages     = {7763--7772}
}

@misc{kim2025seg4diffunveilingopenvocabularysegmentation,
      title={Seg4Diff: Unveiling Open-Vocabulary Segmentation in Text-to-Image Diffusion Transformers}, 
      author={Chaehyun Kim and Heeseong Shin and Eunbeen Hong and Heeji Yoon and Anurag Arnab and Paul Hongsuck Seo and Sunghwan Hong and Seungryong Kim},
      year={2025},
      eprint={2509.18096},
      archivePrefix={arXiv},
      primaryClass={cs.CV},
      url={https://arxiv.org/abs/2509.18096}, 
}

@misc{jin2025matrixmasktrackalignment,
      title={MATRIX: Mask Track Alignment for Interaction-aware Video Generation}, 
      author={Siyoon Jin and Seongchan Kim and Dahyun Chung and Jaeho Lee and Hyunwook Choi and Jisu Nam and Jiyoung Kim and Seungryong Kim},
      year={2025},
      eprint={2510.07310},
      archivePrefix={arXiv},
      primaryClass={cs.CV},
      url={https://arxiv.org/abs/2510.07310}, 
}

@inproceedings{NEURIPS2023_dift,
 author = {Tang, Luming and Jia, Menglin and Wang, Qianqian and Phoo, Cheng Perng and Hariharan, Bharath},
 booktitle = {Advances in Neural Information Processing Systems},
 editor = {A. Oh and T. Naumann and A. Globerson and K. Saenko and M. Hardt and S. Levine},
 pages = {1363--1389},
 publisher = {Curran Associates, Inc.},
 title = {Emergent Correspondence from Image Diffusion},
 url = {https://proceedings.neurips.cc/paper_files/paper/2023/file/0503f5dce343a1d06d16ba103dd52db1-Paper-Conference.pdf},
 volume = {36},
 year = {2023}
}

@inproceedings{NEURIPS2023_sd_dino,
 author = {Zhang, Junyi and Herrmann, Charles and Hur, Junhwa and Polania Cabrera, Luisa and Jampani, Varun and Sun, Deqing and Yang, Ming-Hsuan},
 booktitle = {Advances in Neural Information Processing Systems},
 editor = {A. Oh and T. Naumann and A. Globerson and K. Saenko and M. Hardt and S. Levine},
 pages = {45533--45547},
 publisher = {Curran Associates, Inc.},
 title = {A Tale of Two Features: Stable Diffusion Complements DINO for Zero-Shot Semantic Correspondence},
 url = {https://proceedings.neurips.cc/paper_files/paper/2023/file/8e9bdc23f169a05ea9b72ccef4574551-Paper-Conference.pdf},
 volume = {36},
 year = {2023}
}

@misc{nam2024diffusionmodeldensematching,
      title={Diffusion Model for Dense Matching}, 
      author={Jisu Nam and Gyuseong Lee and Sunwoo Kim and Hyeonsu Kim and Hyoungwon Cho and Seyeon Kim and Seungryong Kim},
      year={2024},
      eprint={2305.19094},
      archivePrefix={arXiv},
      primaryClass={cs.CV},
      url={https://arxiv.org/abs/2305.19094}, 
}

@misc{li2024sd4matchlearningpromptstable,
      title={SD4Match: Learning to Prompt Stable Diffusion Model for Semantic Matching}, 
      author={Xinghui Li and Jingyi Lu and Kai Han and Victor Prisacariu},
      year={2024},
      eprint={2310.17569},
      archivePrefix={arXiv},
      primaryClass={cs.CV},
      url={https://arxiv.org/abs/2310.17569}, 
}

@misc{liu2024understandingcrossselfattentionstable,
      title={Towards Understanding Cross and Self-Attention in Stable Diffusion for Text-Guided Image Editing}, 
      author={Bingyan Liu and Chengyu Wang and Tingfeng Cao and Kui Jia and Jun Huang},
      year={2024},
      eprint={2403.03431},
      archivePrefix={arXiv},
      primaryClass={cs.CV},
      url={https://arxiv.org/abs/2403.03431}, 
}

@misc{blackforestlabs2024flux,
  author       = {Black Forest Labs},
  title        = {Flux: Official inference repository for flux.1 models},
  howpublished = {\url{https://github.com/black-forest-labs/flux}},
  year         = {2024},
  note         = {Accessed: 2024-11-12}
}

@misc{ho2022imagenvideo,
      title={Imagen Video: High Definition Video Generation with Diffusion Models}, 
      author={Jonathan Ho and William Chan and Chitwan Saharia and Jay Whang and Ruiqi Gao and Alexey Gritsenko and Diederik P. Kingma and Ben Poole and Mohammad Norouzi and David J. Fleet and Tim Salimans},
      year={2022},
      eprint={2210.02303},
      archivePrefix={arXiv},
      primaryClass={cs.CV},
      url={https://arxiv.org/abs/2210.02303}, 
}

@InProceedings{Kumari_2023_CVPR_customdiffusion,
    author    = {Kumari, Nupur and Zhang, Bingliang and Zhang, Richard and Shechtman, Eli and Zhu, Jun-Yan},
    title     = {Multi-Concept Customization of Text-to-Image Diffusion},
    booktitle = {Proceedings of the IEEE/CVF Conference on Computer Vision and Pattern Recognition (CVPR)},
    month     = {June},
    year      = {2023},
    pages     = {1931-1941}
}

@inproceedings{NEURIPS2020_ddpm,
 author = {Ho, Jonathan and Jain, Ajay and Abbeel, Pieter},
 booktitle = {Advances in Neural Information Processing Systems},
 editor = {H. Larochelle and M. Ranzato and R. Hadsell and M.F. Balcan and H. Lin},
 pages = {6840--6851},
 publisher = {Curran Associates, Inc.},
 title = {Denoising Diffusion Probabilistic Models},
 url = {https://proceedings.neurips.cc/paper_files/paper/2020/file/4c5bcfec8584af0d967f1ab10179ca4b-Paper.pdf},
 volume = {33},
 year = {2020}
}

@InProceedings{pmlr-v139-radford21a_clip,
  title = 	 {Learning Transferable Visual Models From Natural Language Supervision},
  author =       {Radford, Alec and Kim, Jong Wook and Hallacy, Chris and Ramesh, Aditya and Goh, Gabriel and Agarwal, Sandhini and Sastry, Girish and Askell, Amanda and Mishkin, Pamela and Clark, Jack and Krueger, Gretchen and Sutskever, Ilya},
  booktitle = 	 {Proceedings of the 38th International Conference on Machine Learning},
  pages = 	 {8748--8763},
  year = 	 {2021},
  editor = 	 {Meila, Marina and Zhang, Tong},
  volume = 	 {139},
  series = 	 {Proceedings of Machine Learning Research},
  month = 	 {18--24 Jul},
  publisher =    {PMLR},
  url = 	 {https://proceedings.mlr.press/v139/radford21a.html}
}

@InProceedings{Caron_2021_ICCV_dino,
    author    = {Caron, Mathilde and Touvron, Hugo and Misra, Ishan and J\'egou, Herv\'e and Mairal, Julien and Bojanowski, Piotr and Joulin, Armand},
    title     = {Emerging Properties in Self-Supervised Vision Transformers},
    booktitle = {Proceedings of the IEEE/CVF International Conference on Computer Vision (ICCV)},
    month     = {October},
    year      = {2021},
    pages     = {9650-9660}
}

@misc{song2022_ddim,
      title={Denoising Diffusion Implicit Models}, 
      author={Jiaming Song and Chenlin Meng and Stefano Ermon},
      year={2022},
      eprint={2010.02502},
      archivePrefix={arXiv},
      primaryClass={cs.LG},
      url={https://arxiv.org/abs/2010.02502}, 
}

@misc{wang2023modelscopetexttovideotechnicalreport,
      title={ModelScope Text-to-Video Technical Report}, 
      author={Jiuniu Wang and Hangjie Yuan and Dayou Chen and Yingya Zhang and Xiang Wang and Shiwei Zhang},
      year={2023},
      eprint={2308.06571},
      archivePrefix={arXiv},
      primaryClass={cs.CV},
      url={https://arxiv.org/abs/2308.06571}, 
}

@misc{kingma2022autoencodingvariationalbayes_vae,
      title={Auto-Encoding Variational Bayes}, 
      author={Diederik P Kingma and Max Welling},
      year={2022},
      eprint={1312.6114},
      archivePrefix={arXiv},
      primaryClass={stat.ML},
      url={https://arxiv.org/abs/1312.6114}, 
}

@InProceedings{Rombach_2022_CVPR_ldm,
    author    = {Rombach, Robin and Blattmann, Andreas and Lorenz, Dominik and Esser, Patrick and Ommer, Bj\"orn},
    title     = {High-Resolution Image Synthesis With Latent Diffusion Models},
    booktitle = {Proceedings of the IEEE/CVF Conference on Computer Vision and Pattern Recognition (CVPR)},
    month     = {June},
    year      = {2022},
    pages     = {10684-10695}
}

@misc{podell2023sdxlimprovinglatentdiffusion_sdxl,
      title={SDXL: Improving Latent Diffusion Models for High-Resolution Image Synthesis}, 
      author={Dustin Podell and Zion English and Kyle Lacey and Andreas Blattmann and Tim Dockhorn and Jonas Müller and Joe Penna and Robin Rombach},
      year={2023},
      eprint={2307.01952},
      archivePrefix={arXiv},
      primaryClass={cs.CV},
      url={https://arxiv.org/abs/2307.01952}, 
}


%% file: main_template.bib
@String(CVPR= {IEEE Conf. Comput. Vis. Pattern Recog.})

@String(ICCV= {Int. Conf. Comput. Vis.})

@String(CVPR  = {CVPR})

@String(ICCV  = {ICCV})
